\documentclass[11pt, letterpaper]{nyu}
\usepackage[all]{hypcap}

\usepackage{hyperref}[citecolor=lightblue]
\hypersetup{
    colorlinks = false,
    pdfborder  = {0 0 0}
}

\usepackage{microtype}
\usepackage{graphicx}
\usepackage{subfigure}
\usepackage{booktabs} 
\usepackage{float}
\usepackage{amsmath}
\usepackage{amssymb}
\usepackage{mathtools}
\usepackage{amsthm}
\usepackage{caption}
\usepackage{mathrsfs}
\usepackage{nicefrac}
\usepackage{dsfont}
\usepackage{enumitem}
\usepackage{float}
\usepackage[numbers,square,sort&compress]{natbib}
\usepackage{xspace}
\usepackage[capitalize,noabbrev]{cleveref}
\usepackage{subcaption}
\usepackage{wrapfig}
\usepackage{lipsum}
\usepackage{listings}
\usepackage{amsmath}
\usepackage{amssymb}
\usepackage{mathtools}
\usepackage{amsthm}
\usepackage{bbm}
\usepackage{algpseudocode}
\usepackage{setspace}
\usepackage{color}
\usepackage[skins,theorems]{tcolorbox}
\usepackage{xcolor}
\usepackage{tikz}
\usepackage{multirow}
\usepackage{colortbl}   
\usepackage{array}      

\definecolor{deepblue}{rgb}{0,0,0.5}
\definecolor{deepred}{rgb}{0.6,0,0}
\definecolor{deepgreen}{rgb}{0,0.5,0}

\newcommand\pythonstyle{\lstset{
basicstyle=\ttfamily\footnotesize,
language=Python,
morekeywords={self, clip, exp, mse_loss, uniform_sample, concatenate, logsumexp},              
keywordstyle=\color{deepblue},
emph={MyClass,__init__},          
emphstyle=\color{deepred},    
stringstyle=\color{deepgreen},
frame=single,                         
showstringspaces=false
}}

\lstnewenvironment{python}[1][]
{
\pythonstyle
\lstset{#1}
}
{}

\newcommand\pythoninline[1]{{\pythonstyle\lstinline!#1!}}

\usepackage[textsize=tiny]{todonotes}

\usepackage{crossreftools}
\usepackage{dsfont}
\usepackage{nicefrac}
\usepackage{inconsolata}
\usepackage{algorithm}
\usepackage{amssymb}
\usepackage{csquotes}

\usepackage{wrapfig}

\usepackage{pifont}
\newcommand{\xmark}{\textcolor{red}{\ding{55}}}
\newcommand{\pmark}{$\sim$}

\newcommand{\cmark}{\textcolor{green}{\ding{51}}}
\definecolor{rukagreen}{RGB}{220, 240, 210}

\newcommand{\xxnote}[3]{}
\renewcommand{\xxnote}[3]{}

\newcommand{\website}{\href{https://ruka-hand-v2.github.io/}{\textcolor{blue!50!black}{ruka-hand-v2.github.io}}}

\definecolor{olivegreen}{HTML}{3C8031}

\newcommand{\method}{\textsc{Ruka-v2}}
\newcommand{\ruka}{\textsc{Ruka}}

\title{\method{}: Tendon Driven Open-Source Dexterous Hand with Wrist and Abduction for Robot Learning}
\websitedef{\website}

\reportnumber{}

\author{
    {
    \Authfont 
    Xinqi (Lucas) Liu$^{2*}$\quad 
    Ruoxi Hu$^{1*}$\quad 
    Alejandro Ojeda Olarte$^{1}$\quad 
    Zhuoran Chen$^{2}$\\[0.4em]
    Kenny Ma$^1$\quad 
    Charles Cheng Ji$^1$\quad 
    Lerrel Pinto$^1$\quad 
    Raunaq Bhirangi$^1$\quad 
    Irmak Guzey$^1$
    }
}

\affil{$^1$ New York University, \quad $^2$ New York University Shanghai} 
\affil{$^*$Equal contribution}

\correspondingauthors={%
  \href{mailto:irmakguzey@nyu.edu}{irmakguzey@nyu.edu},\,
  \href{mailto:rh4073@nyu.edu}{rh4073@nyu.edu}. $^*$ denotes equal contribution.}

\begin{abstract}
\vspace{-1em}
\textbf{Abstract: }
Lack of accessible and dexterous robot hardware has been a significant bottleneck to achieving human-level dexterity in robots. Last year, we released \ruka{}, a fully open-sourced, tendon-driven humanoid hand with 11 degrees of freedom — 2 per finger and 3 at the thumb - buildable for under \$1,300. It was one of the first fully open-sourced humanoid hands, and introduced a novel data-driven approach to finger control that captures tendon dynamics within the control system. Despite these contributions, \ruka{} lacked two degrees of freedom essential for closely imitating human behavior: wrist mobility and finger adduction/abduction. In this paper, we introduce \method{}: a fully open-sourced, tendon-driven humanoid hand featuring a decoupled 2-DOF parallel wrist and abduction/adduction at the fingers. The parallel wrist adds smooth, independent flexion/extension and radial/ulnar deviation, enabling manipulation in confined environments such as cabinets. Abduction enables motions such as grasping thin objects, in-hand rotation, and calligraphy. We present the design of \method{} and evaluate it against \ruka{} through user studies on teleoperated tasks, finding a 51.3\% reduction in completion time and a 21.2\% increase in success rate. We further demonstrate its full range of applications for robot learning: bimanual and single-arm teleoperation across 13 dexterous tasks, and autonomous policy learning on 3 tasks. All 3D print files, assembly instructions, controller software, and videos are available at \website{}.

\end{abstract}

\begin{document}
\maketitle

\begin{figure}[h]
    \vskip -0.5em
    \centering
    \includegraphics[width=\linewidth]{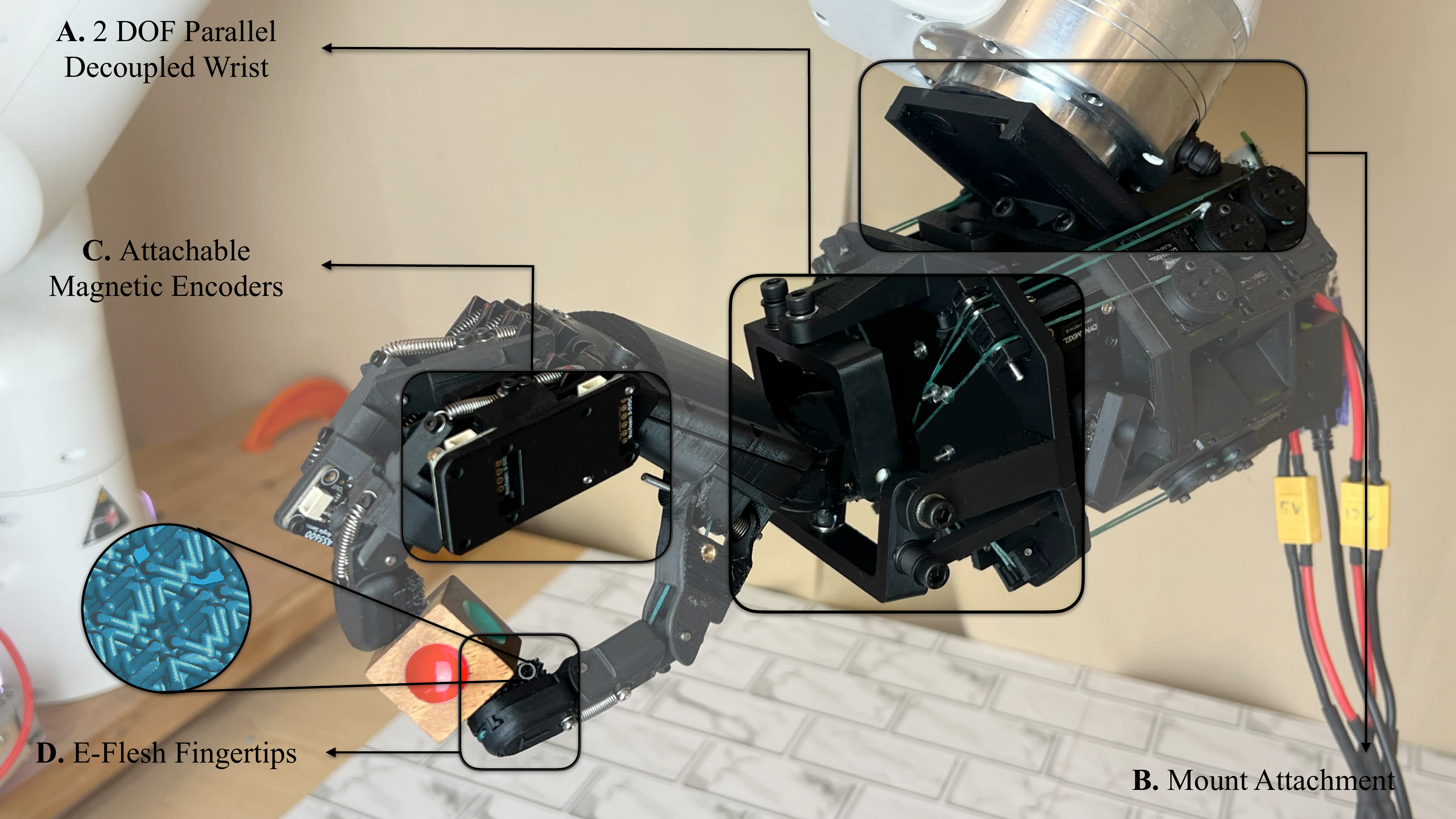}
    \caption{
    (\textbf{A}) \method{} has an integrated 2 DOF decoupled parallel wrist, enabling motion in restricted areas, while also introducing finger abduction/adduction to expand dexterity.
    (\textbf{B}) \method{} has a mount attachment that is located at the side of wrist rather than the bottom, which makes it easier to attach them to table-top manipulators.
    (\textbf{C}) We designed and developed an attachable magnetic encoder to better detect the joint angle accuracy, removing the requirement for expensive motion capture gloves.
    (\textbf{D}) We equipped \method{} with e-flesh fingertips ~\cite{e-flesh} body form factor for softer and compliant grip. This enables optional equipment of tactile sensors on the hand as well.
    Everything is open-source and accessible at our website: \website{}}
    \label{fig:intro}
\end{figure}

\section{Introduction}

Humans complete many tasks in their daily lives effortlessly. While intelligence plays a significant role in this, so does the capability of the human hand. Human hands are compact, possess more than 20 degrees of freedom (DOF), are equipped with a 3 DOF wrist, feature distributed tactile sensing, and have a compliant, soft-bodied form~\cite{wrist-anatomy, human-finger,artificial-wrists, HUME1990Functionalrangeofmotionofthejointsofthehand}. Inspired by these properties, many frameworks in the field have sought to replicate this dexterity in robotic systems. While the community has made significant progress on the algorithmic front~\cite{guzey2025dexteritysmartlensesmultifingered, zhao2023learningfinegrainedbimanualmanipulation, guzey2023dexteritytouchselfsupervisedpretraining, chi2024diffusionpolicy, haldar2025pointpolicyunifyingobservations, cap}, hardware remains a major bottleneck. This gap has pushed research labs and commercial companies alike to improve robotic hand designs with the goal of matching human-level dexterity~\cite{wuji, sharpa, shadow, inspire, x-hand, orca, zorin2025ruka, adapt, rbo3, act-hand}. 

The most simplistic hand design is to attach an actuator directly to each finger segment, creating an exact mapping between joint values and motors. The Allegro~\cite{allegro} and open-sourced LEAP~\cite{leap} hands are great examples of this approach. While both have been widely used in the community, they are typically bulky, heavy, and require significant algorithmic effort to close the human-to-robot embodiment gap~\cite{hudor, spider, dexmachina, anyteleop} in order to complete dexterous autonomous tasks. Recently, much smaller direct-driven hands have been introduced that use smaller actuators attached to the fingers, maintaining a form factor similar to a human hand~\cite{wuji, sharpa}. While these hands offer great dexterity and accuracy, they either overheat fairly quickly (within 30 minutes of use)~\cite{wuji} or are priced very high — around \$50,000~\cite{sharpa} — limiting their accessibility.

To make these hands more compact, there has been a significant amount of past work focusing on tendon-driven hand design~\cite{zorin2025ruka, dex-hand, inmoov, adapt, act-hand, rbo3, orca, shadow}. This design typically moves all actuators outside of the hand, usually toward the wrist, and connects them to the hand joints via a rope-like mechanism that imitates a "tendon." The first commercial example of this is the Shadow Hand~\cite{shadow}, which demonstrated great capabilities over the years, but these hands are very expensive — priced at more than \$100,000 — and are difficult to repair due to their complicated design. To make this more accessible, a growing number of tendon-driven hands have emerged from research institutions. Some of these hands are limited in dexterity, difficult to use, or offer a lower number of DOF~\cite{tetheria, inmoov}; others, while dexterous, are either not fully open-source — preventing modification to printable files~\cite{orca} — or lack sufficient documentation, causing difficulties in reproduction~\cite{leapv2, dex-hand, adapt, act-hand, rbo3}. Notably, despite humans relying heavily on wrist motion to accomplish most daily tasks, the majority of these designs focus on the hand itself instead of the wrist. 

Building on our prior work \textsc{Ruka}, we introduce \method{}: a humanoid, tendon-driven hand featuring a decoupled parallel 2 DOF wrist and abduction/adduction capabilities across the fingers. We open-source all 3D printable files alongside documentation supported by instructional videos, teleoperate \method{} to complete 10 single arm and 3 bimanual dexterous tasks including writing, and demonstrate learned autonomous policies on 3 tasks, showcasing its utility for robot learning. We also conduct user experience experiments comparing \method{} to \ruka{} by asking users to teleoperate 3 different tasks while recording their completion times and success rates. We list our contributions as follows:
\begin{enumerate}
    \item We demonstrate a 51.3\% reduction in completion time and a 21.2\% increase in success rate over \ruka{} on user experience, highlighting the need for capable wrist actuation and abduction/adduction. We further elaborate on these differences in Fig.~\ref{fig:v1-comparison}. (Section~\ref{sec:v1-comparison})
    \item We elaborate on a set of design principles that make hands easier to use in robot learning, such as wrist attachments and soft fingertips. 
    \item We design attachable magnetic sensors per joint, that can be fitted to any \method{} (and \ruka{} as well) to calibrate and detect joint angle errors across all tendon-driven joints. These attachments are also open-sourced. We evaluate the accuracy of our control mechanism of \method{} in Section~\ref{sec:controller_accuracy} and describe the sensor mechanism in detail in Section~\ref{sec:magnetic_encoder}.
\end{enumerate}
Collectively, we believe that \method{} will serve as a valuable tool for dexterous manipulation research, both for algorithmic development and, with modifiable CAD files, for hardware research as well. We provide a thorough comparison of \method{} to other listed robot  hardware in Table~\ref{tab:dex_hand_comparison}. All hardware and software components, along with robot videos, are available at: \website{}.


\begin{table*}[t]
\centering
\setlength{\tabcolsep}{4pt}
\renewcommand{\arraystretch}{1.1}
\footnotesize

\caption{%
  Comparison of dexterous robotic hands.
  \textbf{Open-source}: \cmark\,=\,hardware \& software fully public;
  $\sim$\,=\,partial; \xmark\,=\,proprietary.
  \textbf{Act.}: T\,=\,tendon; DD\,=\,direct-drive.
  \textbf{Tactile}: opt.\,=\,optional add-on.
  $^\dagger$HW files public; SW closed.
  Costs are approx.\ USD list or material prices;
  {--}\,=\,not publicly documented.
  $^\ddagger$Price not officially disclosed; figure reported by third-party sources only.
  $^\S$Wuji Hand thermal throttling observed after ${\sim}$40 minutes of continuous operation in internal testing.
}
\label{tab:dex_hand_comparison}

\begin{tabular}{@{} l *{7}{>{\centering\arraybackslash}p{1.63cm}} @{}}
\toprule

\multirow{2}{*}{\textbf{Metric}}
  & \includegraphics[width=1.4cm,height=1.5cm,keepaspectratio]{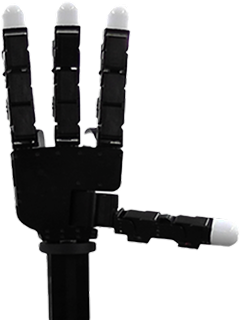}
  & \includegraphics[width=1.4cm,height=1.5cm,keepaspectratio]{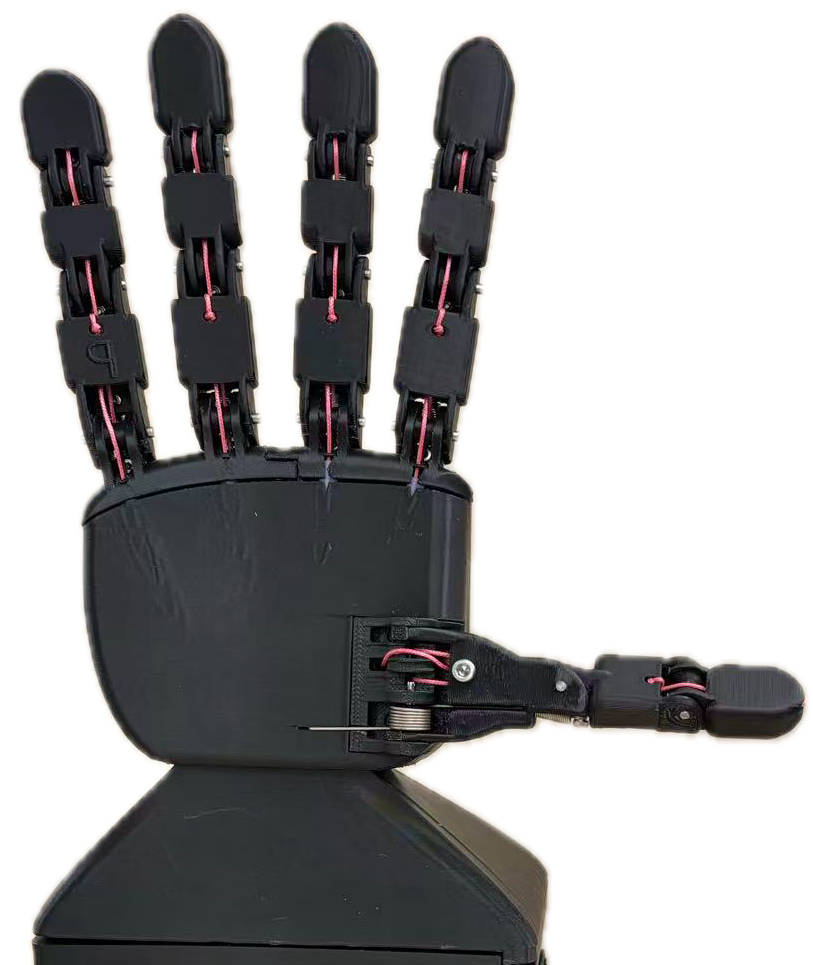}
  & \includegraphics[width=1.4cm,height=1.5cm,keepaspectratio]{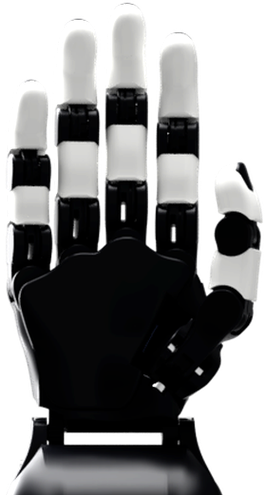}
  & \includegraphics[width=1.4cm,height=1.5cm,keepaspectratio]{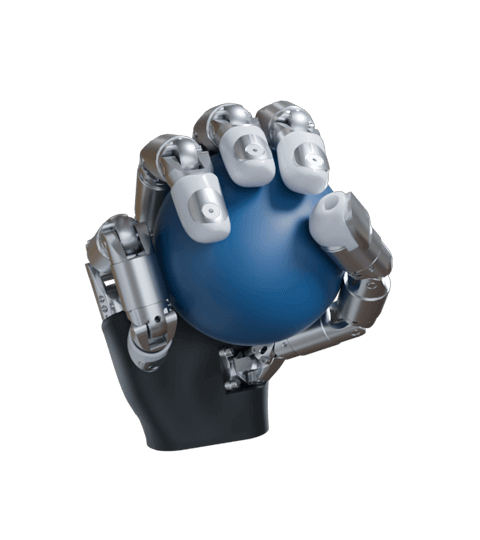}
  & \includegraphics[width=1.4cm,height=1.5cm,keepaspectratio]{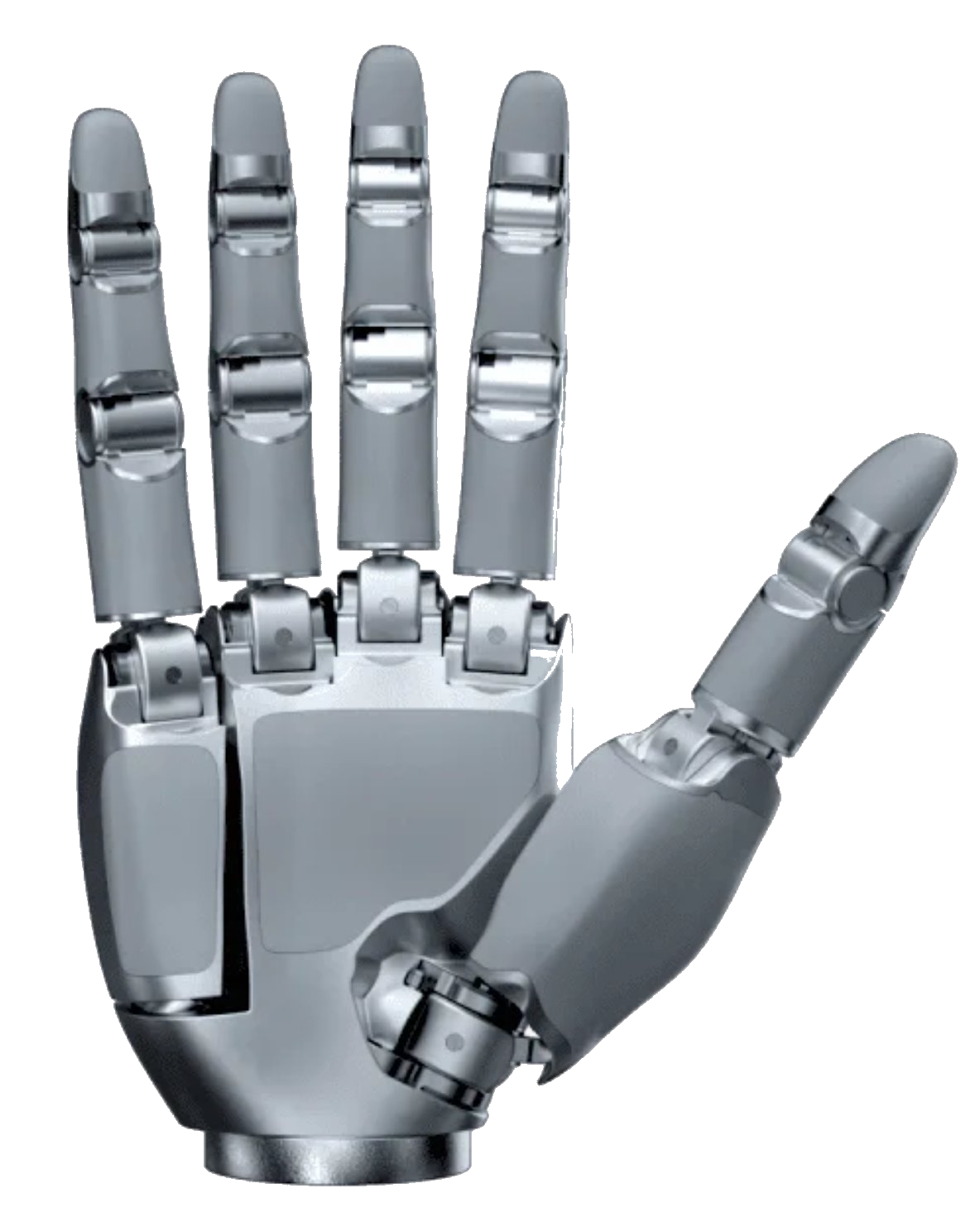}
  & \includegraphics[width=1.4cm,height=1.5cm,keepaspectratio]{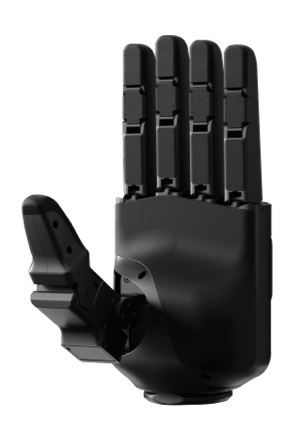}
  & \cellcolor{rukagreen}\includegraphics[width=1.4cm,height=1.5cm,keepaspectratio]{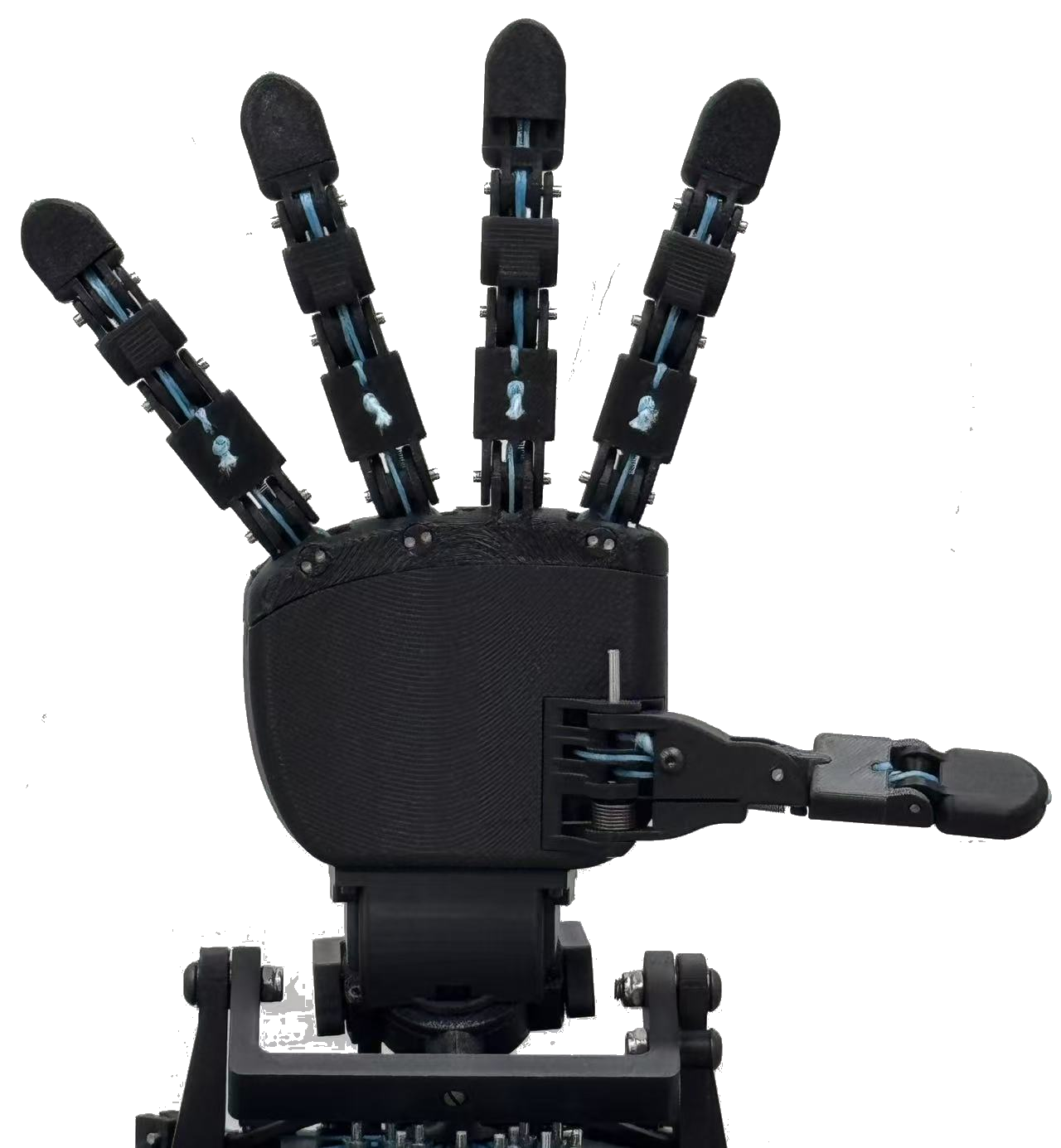} \\[2pt]

  & \textbf{Allegro V4}
  & \textbf{Ruka v1}
  & \textbf{ORCA}
  & \textbf{Wuji Hand}
  & \textbf{Sharpa Wave}
  & \textbf{Unitree Dex5}
  & \cellcolor{rukagreen}\textbf{\method{} (Ours)} \\

\midrule

Open-source
  & \pmark$^\dagger$ & \cmark   & \cmark
  & \xmark           & \xmark   & \xmark
  & \cellcolor{rukagreen}\cmark \\

Cost (USD)
  & ${\sim}$\$16K   & ${\sim}$\$1.3K & $\sim$\$3.5K
  & ${\sim}$\$5.5K  & ${\sim}$\$50K$^\ddagger$ & ${\sim}$\$25K
  & \cellcolor{rukagreen} $\sim$\$1.5K \\

\midrule

Tot.\,/\,Act.\ DOF
  & 16\,/\,16 & 15\,/\,11 & 17\,/\,17
  & 20\,/\,20 & 22\,/\,22 & 21\,/\,
  & \cellcolor{rukagreen}20\,/\,16 \\

Wrist DOF
  & 0  & 0  & 1  & 0  & 0  & 0
  & \cellcolor{rukagreen} 2 \\

Actuation
  & DD & T  & T  & DD & DD & DD
  & \cellcolor{rukagreen} T \\

Hand size
  & ${\sim}$30\%larger & Human & Human
  & Human & Human & Human
  & \cellcolor{rukagreen} Human \\

\midrule

Tactile
  & opt. & \xmark & \cmark\,(FSR)
  & \xmark & \cmark\,(DTA) & \cmark\,(opt.)
  & \cellcolor{rukagreen}\cmark\,(opt.)\\

\midrule

Durability
  & High & High & $>$10K cyc.
  & ${\sim}$40\,min$^\S$ & $>$2.5M cyc. & High
  & \cellcolor{rukagreen} $>$5Hours \\

\bottomrule
\end{tabular}
\end{table*}
\section{Hardware Design}
\label{sec:hardware_design}

\begin{figure}[t]
    \centering
    \includegraphics[width=\linewidth]{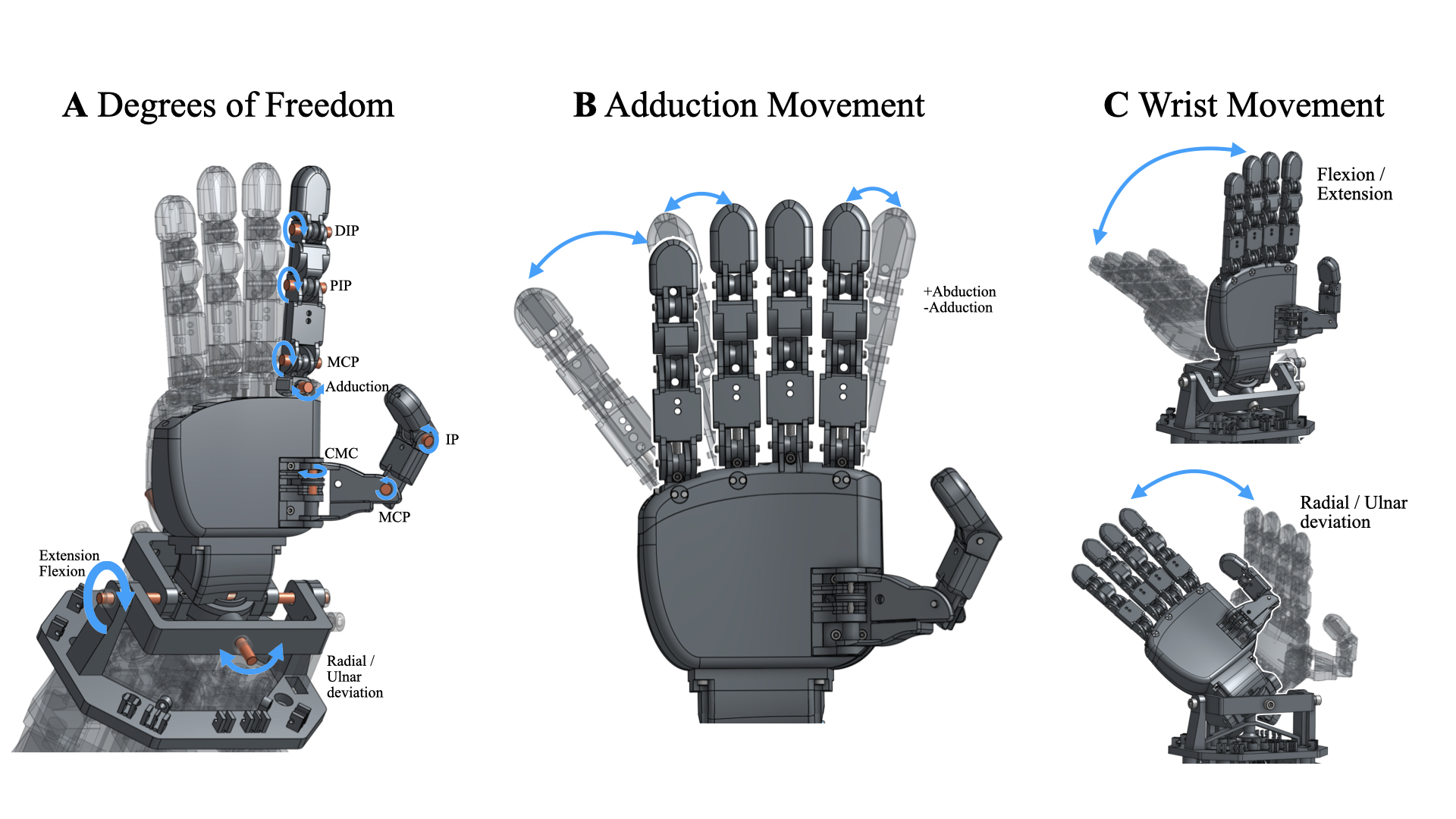}
    \caption{\textbf{\method{} hardware overview.}
    (\textbf{A}) \method{} features 18 degrees of freedom across the fingers and thumb,
    with joints at the DIP, PIP, MCP, and Adduction axes on each finger, and IP, MCP,
    and CMC joints on the thumb. An additional 2~DOF wrist provides flexion/extension
    and radial/ulnar deviation.
    (\textbf{B}) The independent knuckle module enables abduction/adduction at the MCP
    joints, allowing the fingers to splay and converge laterally. The middle finger
    remains fixed as a structural reference.
    (\textbf{C}) The 2~DOF wrist module supports near-human range of motion: flexion/extension
    (top) and radial/ulnar deviation (bottom), actuated independently via decoupled
    parallel linkages.}
    \label{fig:hardware_overview}
\end{figure}

Building on \ruka{}, \method{} introduces two hardware capabilities critical for human-like manipulation: (i) a decoupled 2-DOF parallel wrist and (ii) controlled finger abduction/adduction. We also present an optional DIP/PIP joint coupling system and a detachable magnetic encoder array mountable across all joints of \ruka{} and \method{} for calibration and joint angle sensing. All major structural components are 3D printed; bearings, fasteners, and springs are off-the-shelf parts. Actuators are placed proximally in the forearm to reduce distal inertia and simplify maintenance. Tendons route through the wrist and palm to each finger via guides that prevent sharp bends and reduce tension variation across wrist poses. In this section, we describe each of these components in detail and provide illustration of overall functionality of \method{} in Figure \ref{fig:hardware_overview}. 

\subsection{Design Principles}

\method{} is designed to maximize functionality by expanding the degrees of freedom offered by \ruka{}, while preserving accessibility and reliability. We detail these design principles in this section and summarize them with a comparative analysis against other hands in Table \ref{tab:dex_hand_comparison}.

\paragraph{Range of Motion and Speed}

To best replicate human hand behavior, the hardware must support a range of motion comparable to that of a human hand. Table \ref{tab:rom} compares the per-finger, adduction, and wrist range of motion of \method{} against human values. \method{} closely matches human range-of-motion requirements across all joints.

\begin{table}[h!]
    \centering
    \caption{ The range of motion ($\uparrow$) of each joint in \method{} in degrees compared to the human range of motion \cite{HUME1990Functionalrangeofmotionofthejointsofthehand}. \cmark = more or equal to human; $\sim$ = DOF exists but range of motion not reached; \xmark = DOF does not exist.}
    \begin{tabular}{lcccc}
        \toprule
        \textbf{\raggedright Type} & \textbf{\raggedright Requirement} & \textbf{Human} & \textbf{\method{}} & \textbf{Pass} \\ 
        \midrule
        & Distal Interphalangeal (DIP) & 85\textdegree & 120\textdegree & \cmark \\
        \textbf{Fingers} & Proximal Interphalangeal (PIP) & 105\textdegree & 120\textdegree & \cmark \\ 
        & Metacarpophalangeal (MCP, Finger) & 85\textdegree & 140\textdegree & \cmark \\ 
        \midrule
        & Interphalangeal (IP) & 80\textdegree & 120\textdegree & \cmark \\ 
        \textbf{Thumb} & Metacarpophalangeal (MCP, Thumb) & 56\textdegree & 90\textdegree & \cmark \\ 
        & Carpometacarpal (CMC) & $--$ & 170\textdegree & \cmark \\  
        \midrule 
        & Index & 20-25\textdegree & 20\textdegree & \cmark \\
        & Middle & 10-12\textdegree & Fixed & \xmark \\
        \textbf{Adduction/Abduction} & Ring & 20-25\textdegree & 23\textdegree & \cmark \\
        & Pinky & 30-35\textdegree & 45\textdegree & \cmark \\ 
        \midrule 
        & Flexion & 80-90\textdegree & 45\textdegree & $\sim$ \\
        & Extension & 70\textdegree & 30\textdegree & $\sim$ \\ 
        \textbf{Wrist} & Radial deviation & 20\textdegree & 35\textdegree & \cmark \\
        & Ulnar deviation & 30\textdegree & 35\textdegree & \cmark \\
        \bottomrule
    \end{tabular}
    
    \label{tab:rom}
\end{table}

\paragraph{Accessibility} For \method{} to serve as a reliable research tool, it must be reproducible across institutions with varying resources. This demands both thorough documentation and low cost. We open-source all design files, including modifiable CAD models, enabling not just reproduction but also further hardware research and customization. Build instructions are documented at every step, supplemented by video tutorials, and the only specialized equipment required is a 3D printer — all remaining components are off-the-shelf products that can be purchased directly. This removes the dependency on materials such as silicone, which typically requires specialized hardware and substantial expertise to work with reliably. Notably, \method{} achieves a soft form factor on the fingertips — a property that usually necessitates silicone casting — by using E-flesh~\cite{e-flesh} fingertip attachments, entirely through 3D-printed components, maintaining accessibility without sacrificing compliance. The total material cost of \method{} is under \$1,500, making it accessible to a wide range of research labs.

\paragraph{Reliability} For hardware to be suitable for robot learning, it must remain consistent and durable over extended experimental periods. This makes thermal stability across long runs and ease of repair two critical factors. We evaluate the thermal stability of \method{} under extended operation in Section \ref{sec:experiments}, where we showcase promising results. Additionally, the hand's open-source design enables quick, in-house repairs, resulting in significantly less downtime compared to commercial robot hands.

\subsection{Wrist Kinematics}
\label{sec:wrist_module}

\begin{figure}[h]
    \centering
    \includegraphics[height=8.2cm, width=\linewidth]{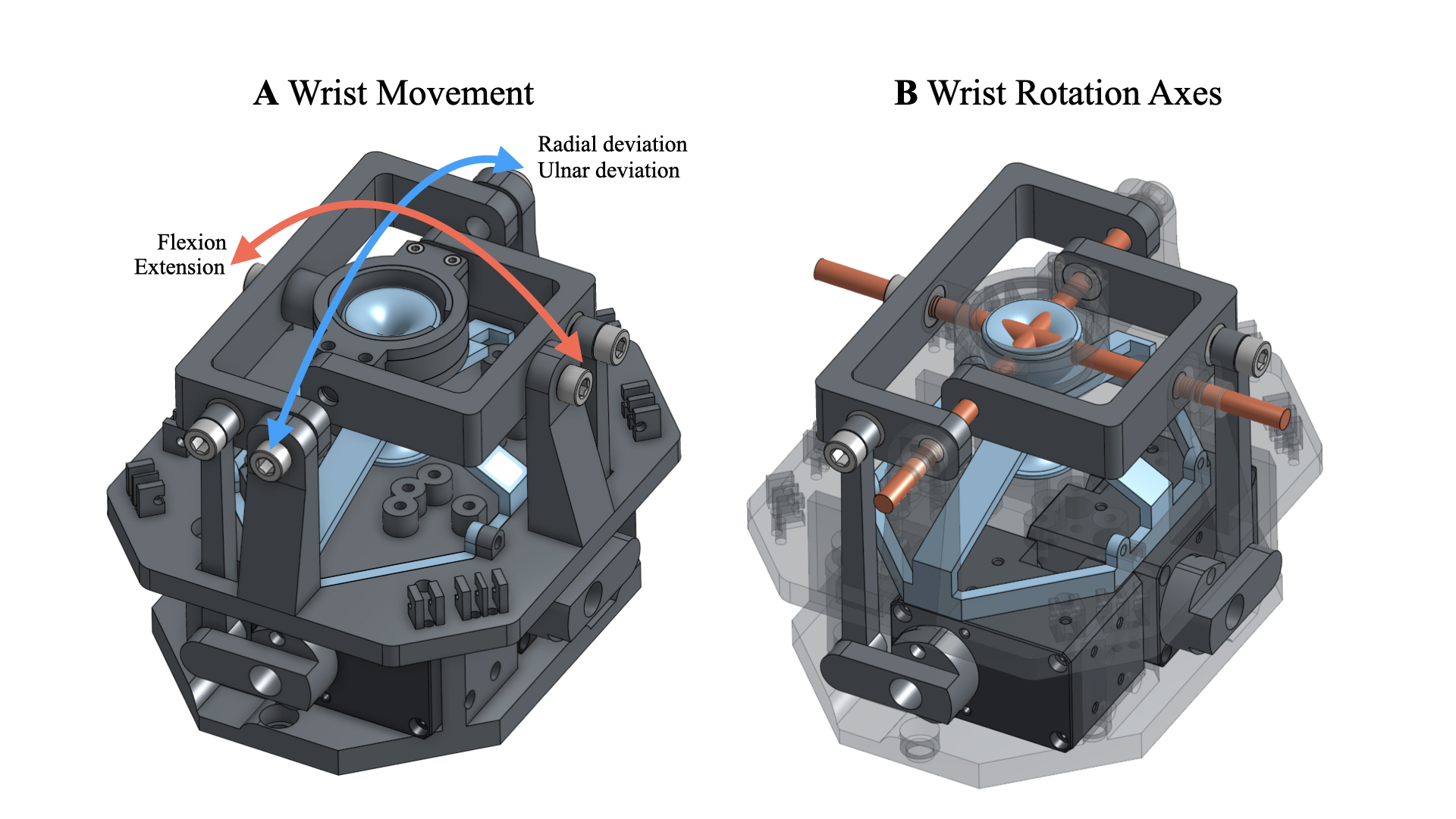}
    \caption{\textbf{2 DOF wrist kinematics.} (\textbf{A}) The wrist provides two independent
    degrees of freedom: flexion/extension (red arrow) and radial/ulnar deviation (blue arrow),
    both actuated through independent linkage chains meeting at a common pivot point.
    (\textbf{B}) The intersecting rotation axes share a single geometric center defined by a
    passive spherical ball joint. Each DOF is driven by a dedicated forearm motor via a
    rectangular linkage, decoupling the two axes and minimizing cross-axis coupling during
    wrist motion.}
    \label{fig:wrist_axes}
\end{figure}

Since all finger tendons must route through the wrist, designing a 2-DOF wrist that does not disturb finger motion is non-trivial. In \method{}, the wrist provides flexion/extension and radial/ulnar deviation while maintaining predictable tendon geometry, making it well-suited for tendon-driven actuation in constrained environments. The wrist components are illustrated in Figure \ref{fig:wrist_axes}.

\paragraph{Decoupled parallel wrist design}
To best mimic a human wrist, we adopt a decoupled parallel wrist mechanism similar to~\cite{dexwrist}, in which flexion/extension and radial/ulnar deviation are actuated independently. The wrist motors are mounted just below the wrist and connected to rectangular linkages that surround it, whose rotation axes meet at a single geometric pivot point. This minimizes translational motion of the palm during wrist rotation and simplifies modeling and calibration. Mechanically, a passive spherical ball joint defines the common center of rotation, while each DOF is driven by an independent linkage chain actuated by a forearm motor. Decoupling pivot definition (ball joint) from actuation (linkages) yields smooth motion and reduces cross-axis coupling compared to offset-axis or serial wrist designs.

\paragraph{Tendon routing through the rotation center}
A through-hole in the spherical joint allows finger tendons to pass near the wrist rotation center. A bearing-supported routing plate then redirects tendons outward into the forearm, preventing sharp bends and reducing tendon entanglement during large wrist rotations. This co-design limits wrist-induced variation in effective tendon length, improving controllability for both teleoperation and learned policies.

\subsection{Finger Abduction/Adduction Module}
\label{sec:abduction}
Human grasping frequently requires adjusting finger spacing (e.g., pinching thin objects, conforming to irregular geometry, and stabilizing objects during in-hand reorientation). To enable these behaviors,
\method{} adds controlled abduction/adduction at the MCP joints.

\paragraph{Independent knuckle module}
The palm incorporates independent knuckle modules that house both the primary MCP flexion axis and a secondary abduction/adduction axis, enabling lateral rotation at the finger base. The middle finger remains fixed to provide a stable geometric reference for alignment and a stiff structural backbone.
\paragraph{Single-tendon actuation with spring return}
Each abduction/adduction joint is driven by a dedicated tendon routed from a forearm motor, while a small extension spring provides the torque to restore the finger back to neutral when the tendon is relaxed. This configuration adds compliance during contact, reduces actuator requirements for bidirectional control, and enforces a well-defined default posture.

\begin{figure}[h]
    \centering
    \includegraphics[width=0.8\linewidth]{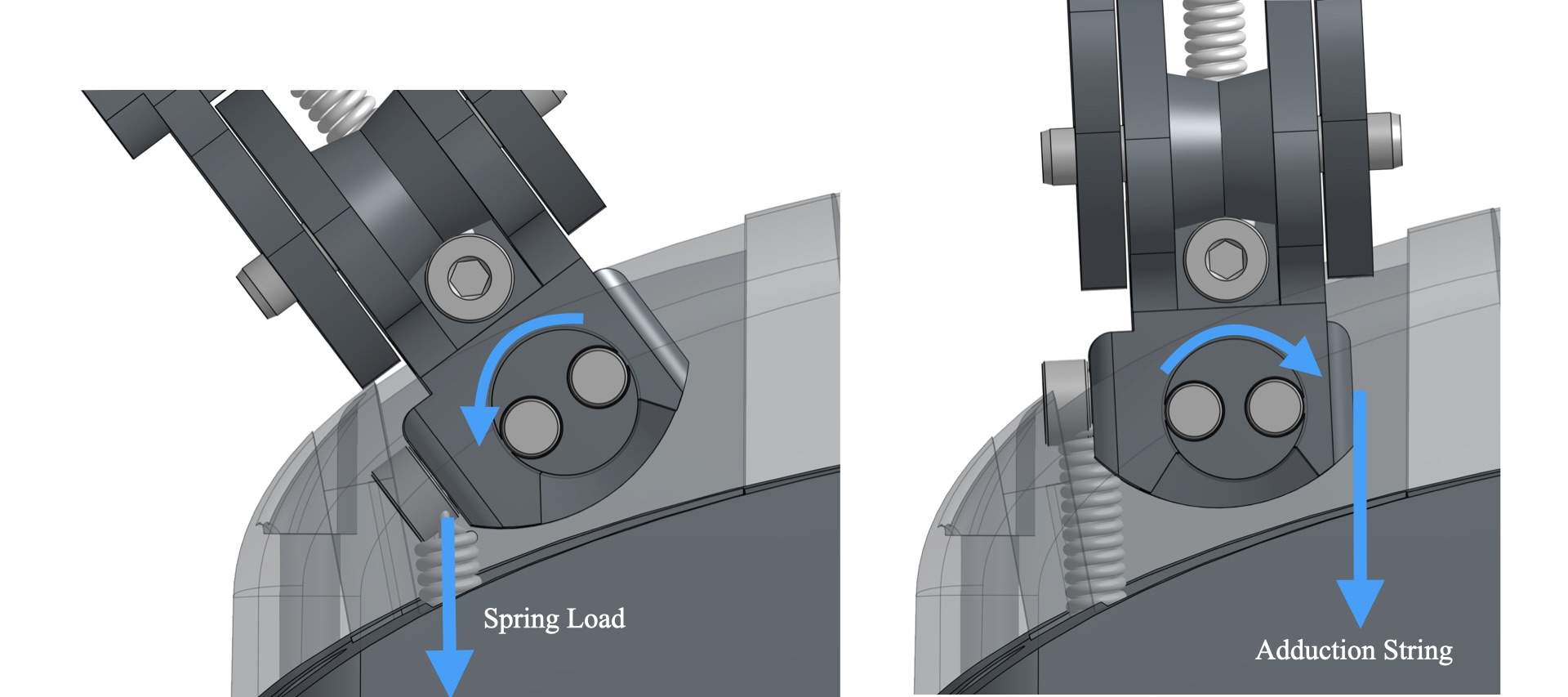}
    \caption{\textbf{Finger abduction/adduction mechanism.} (\textbf{Left}) In the abducted state,
    the extension spring pulls the finger laterally outward, providing passive compliance and a
    well-defined neutral posture when the tendon is slack.
    (\textbf{Right}) Tendon-driven adduction: the adduction string is pulled by a forearm motor,
    rotating the knuckle module inward against the spring force.
    Blue arrows indicate the direction of applied force.}
    \label{fig:abduction_module}
\end{figure}

\subsection{Optional DIP/PIP Coupling}
\label{sec:coupling}

\begin{figure}[h]
    \centering
    \includegraphics[height=3.7cm,width=0.5\linewidth]{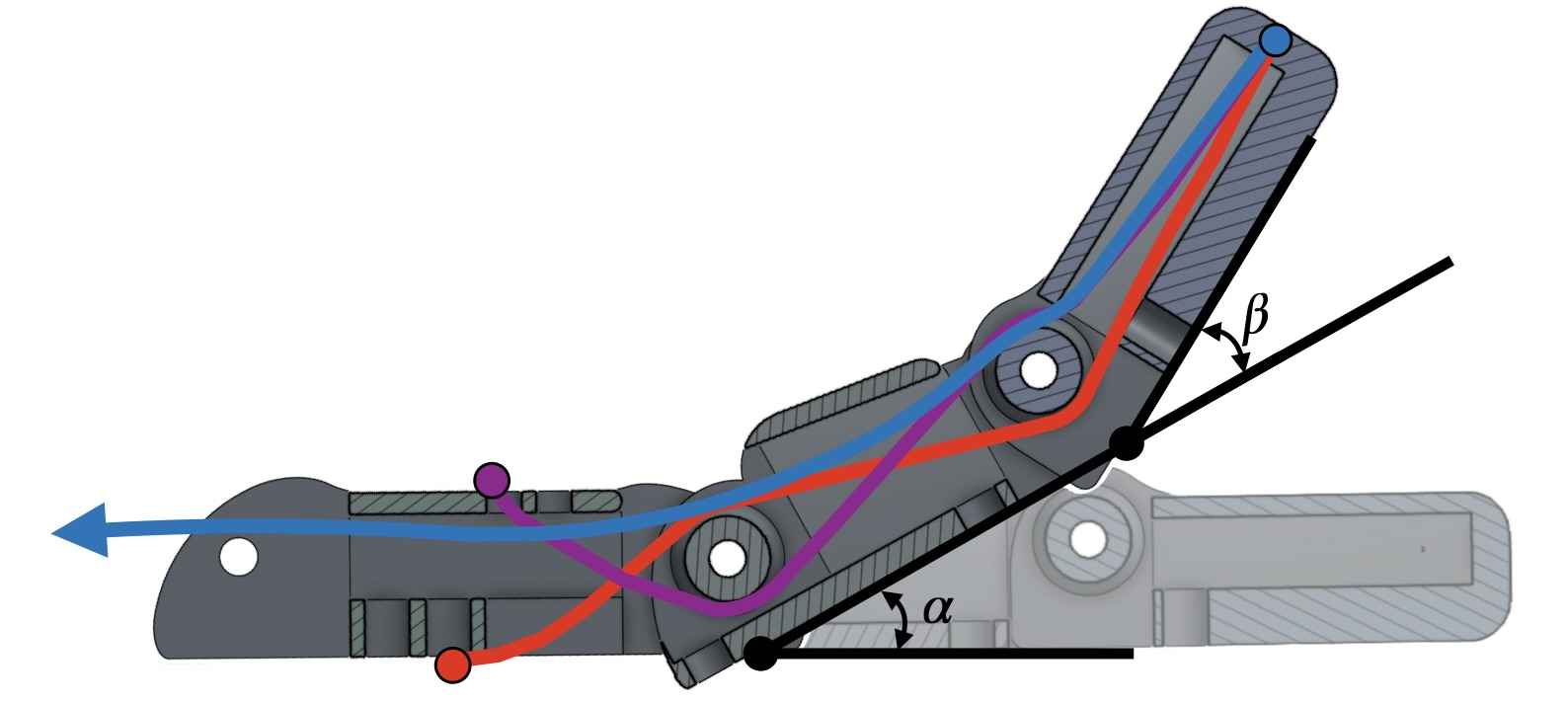}
    \caption{\textbf{DIP--PIP coupling mechanism.} Fixed-length coupling strings (Red \& Purple) routed through internal channels, enforce a deterministic relationship between PIP and DIP motion ($\alpha \approx \beta$).}
    \label{fig:coupling_routing}
\end{figure}

In both \method{} and \ruka{}, the PIP and DIP joints are actuated by a single motor, coupling their motion. There are two approaches to designing this coupling. In \ruka{}, although both joints share a single motor, no mechanism actively enforces equal joint displacement — meaning that due to friction, slack, and wear, the relative motion between the two joints varies across different hand builds, complicating the joint-to-motor mapping. However, this also introduces passive compliance: under external forces, each joint can deflect independently, adding flexibility to the overall motion. In \method{}, we provide an optional rigid coupling mechanism as an add-on design, with full instructions, that enforces synchronized PIP/DIP motion — improving repeatability and simplifying control at the cost of passive compliance. We describe the design and assembly of this coupling in detail below and provide an illustration in Figure \ref{fig:coupling_routing}.

\method{} uses fixed-length coupling strings that enforce a deterministic relationship between PIP and DIP angles. The strings route through enlarged internal channels and anchor to the proximal and distal segments such that DIP flexion follows PIP flexion in a consistent manner across repeated cycles. By mechanically enforcing the coupling, this approach reduces sensitivity to tendon friction and slack and yields more repeatable curling behavior.

\subsection{Attachable Magnetic Encoders}
\label{sec:magnetic_encoder}

\begin{figure}[H]
    \centering
    \includegraphics[height=5cm, width=0.9\linewidth]{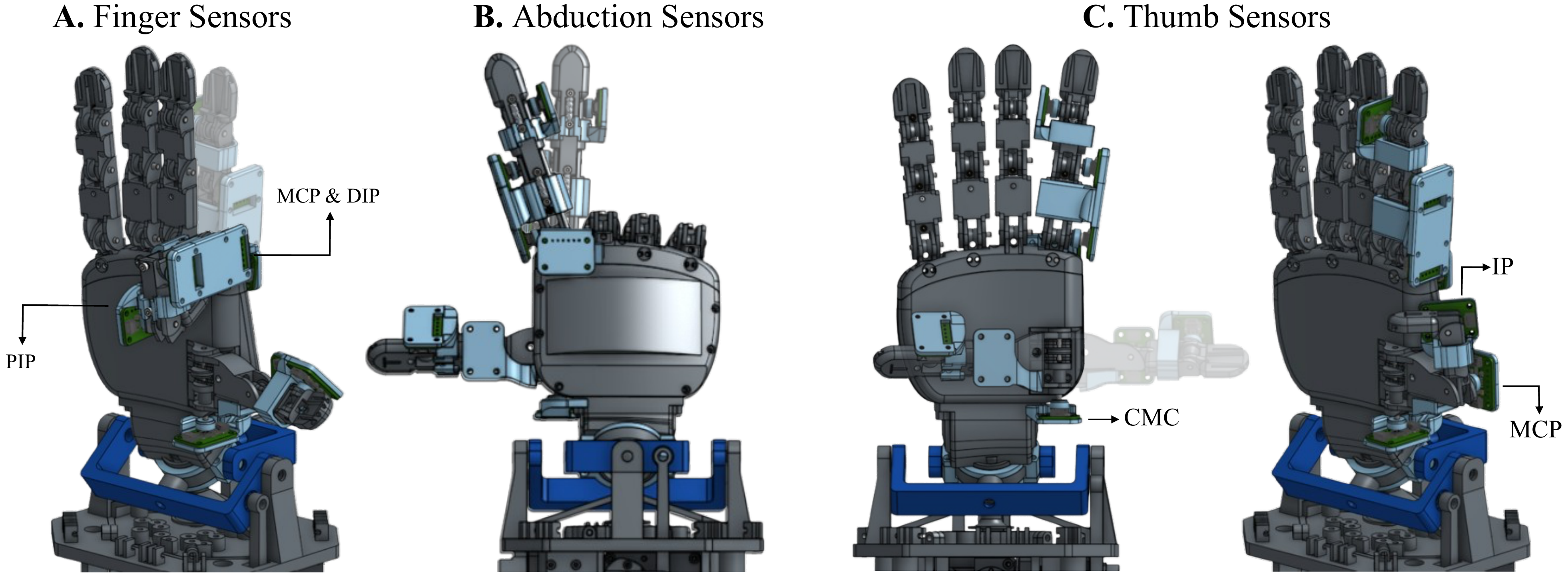}
    \caption{\textbf{Magnetic encoder motion.}  (\textbf{A}) The finger sensors hold two pieces that read the MCP, PIP and DIP of the index finger.  (\textbf{B}) The abduction sensor is placed in the back to avoid interference with grasping tasks with the closed hand.  (\textbf{C}) For the thumb, the CMC sensor is placed in the bottom section of the palm, and the other two are attached to the MCP and IP sections of the thumb. }
    \label{fig:sensors}
\end{figure}

To evaluate the accuracy of the perceived joint angles, a set of detachable magnetic angle sensors were designed to measure the current angle of the joint. The set uses AS5600 magnetic angle sensors, which provide 12-bit resolution to the range of motion. The set is composed by rare earth magnet holders and 6 parts that are attachable to the index and thumb fingers, 3 for each finger, as shown in Figure \ref{fig:sensors}. 

The magnet holders are press-fitted in the dowel pin of the joint. The sensor structures are attached on the DIP and MCP sections for the thumb and index fingers and on the bottom side section of the wrist for the index abduction/adduction joint. All components of the encoders set are secured using screws that already are part of the original hand design, thereby eliminating the need for any additional fastening elements. An ESP32 QTPy microcontroller interfaces the sensor readings over an I2C multiplexer board and the host computer for calibration and data collection.

\section{Applications of \method{}}

We hope that \method{} will serve as a suitable research tool for robot learning. To evaluate its usability, we implement a joint-space controller, teleoperate \method{} across 10 single-arm tasks and 3 bimanual tasks, and train robot policies for 3 tasks that perform reliably across multiple runs. We describe each of these components in detail below. Software stack including the controller mechanism and the calibration is open-sourced and can be found on our website: \website{}. 

\subsection{Controller}
\label{sec:controller}

For tendon-driven hands, while the retargeting problem — mapping human hand poses to robot joint angles — is eased by the morphological similarity between the robot hand and human, the mapping between joint angles and motor values remains non-trivial. \ruka{} introduced a data-driven approach to address both problems jointly. While effective, this relied on expensive motion capture gloves, hindering reproducibility. In \method{}, we decouple these two problems and treat retargeting and joint-to-motor mapping separately.

\paragraph{Linear modeling of joint-to-motor relationship} Similar to~\cite{orca}, we define a linear mapping between joint angles $\theta \in [\theta_{\min}, \theta_{\max}]$ and motor positions $p \in [p_{\min}, p_{\max}]$. Desired joint angles $\theta$ are translated to motor positions $p$ via:
$$p = p_{\min}+ c \cdot \frac{\theta}{\theta_{\max}-\theta_{\min}} \cdot (p_{\max}-p_{\min})$$
A configurable scaling factor $c$ compensates for any additional control effort required due to internal tendon friction and the weight of the finger, or joint angle underestimation in hand detection frameworks for any given joint, most commonly observed in the MCP joints.

To determine motor and joint limits, we perform per-motor calibration. The minimum motor position $p_{\min}$ corresponds to the fully extended joint state and is defined as the point of initial tendon tensioning. The maximum position $p_{\max}$ is set by the joint's mechanical limit in the fully curled position. This calibration ensures that the linear mapping accurately reflects \method{}'s physical range of motion. This calibration is an automated procedure and can be done by the calibration script provided on our codebase.

\paragraph{Retargeting human hand pose to \method{}} We adopt the retargeting module from AnyTeleop~\cite{anyteleop} to map human hand poses to \method{} joint angles. Raw 3D hand keypoints are extracted from a video stream and transformed into \method{}'s base coordinate frame. We then apply the vector-based Dex Retargeting module from AnyTeleop, which optimizes the robot's joint angles $\theta$ to align the link vectors of the kinematic chain of \method{} with the corresponding human finger directions, rather than relying on traditional inverse kinematics to match fingertip positions alone.

\begin{figure}[H]
    \centering
    \includegraphics[width=0.95\linewidth]{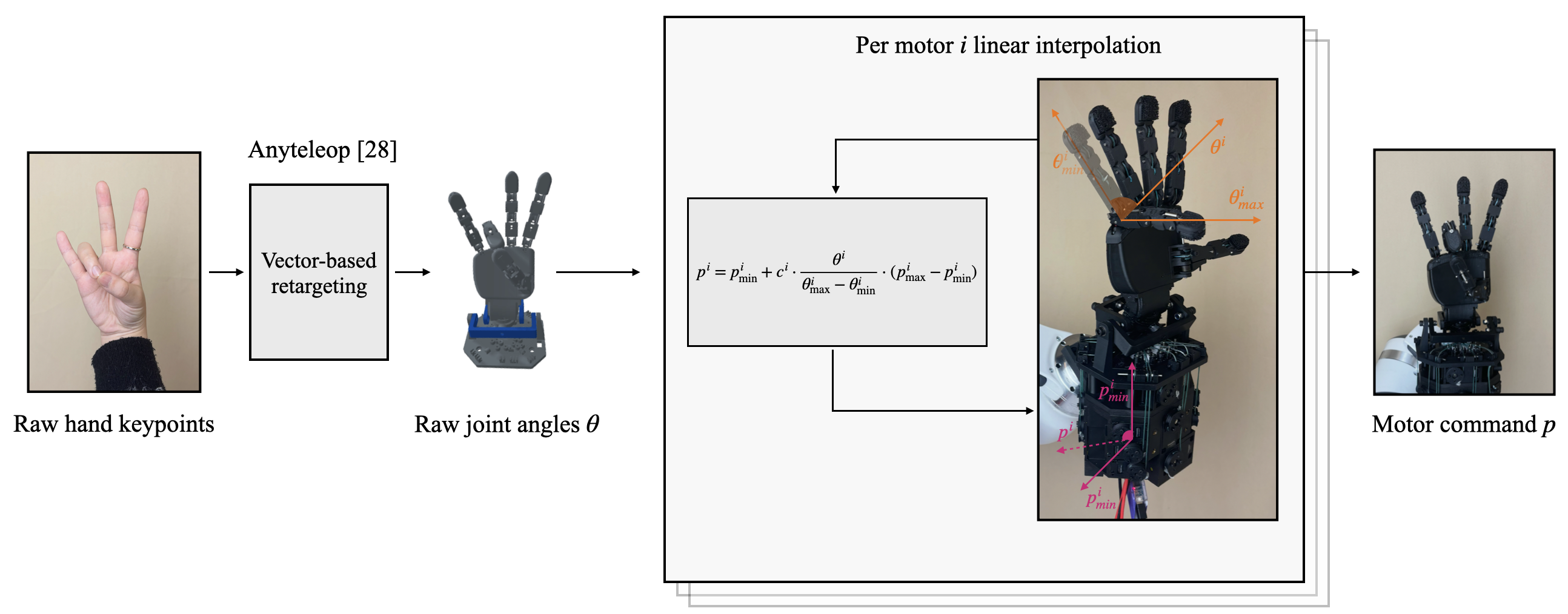}
    \caption{Illustration of the linear controller}
    \label{fig:controller}
\end{figure}

\subsection{Teleoperation}
\label{sec:teleoperation}
\begin{figure}[p]
    \centering
    \includegraphics[width=0.98\linewidth]{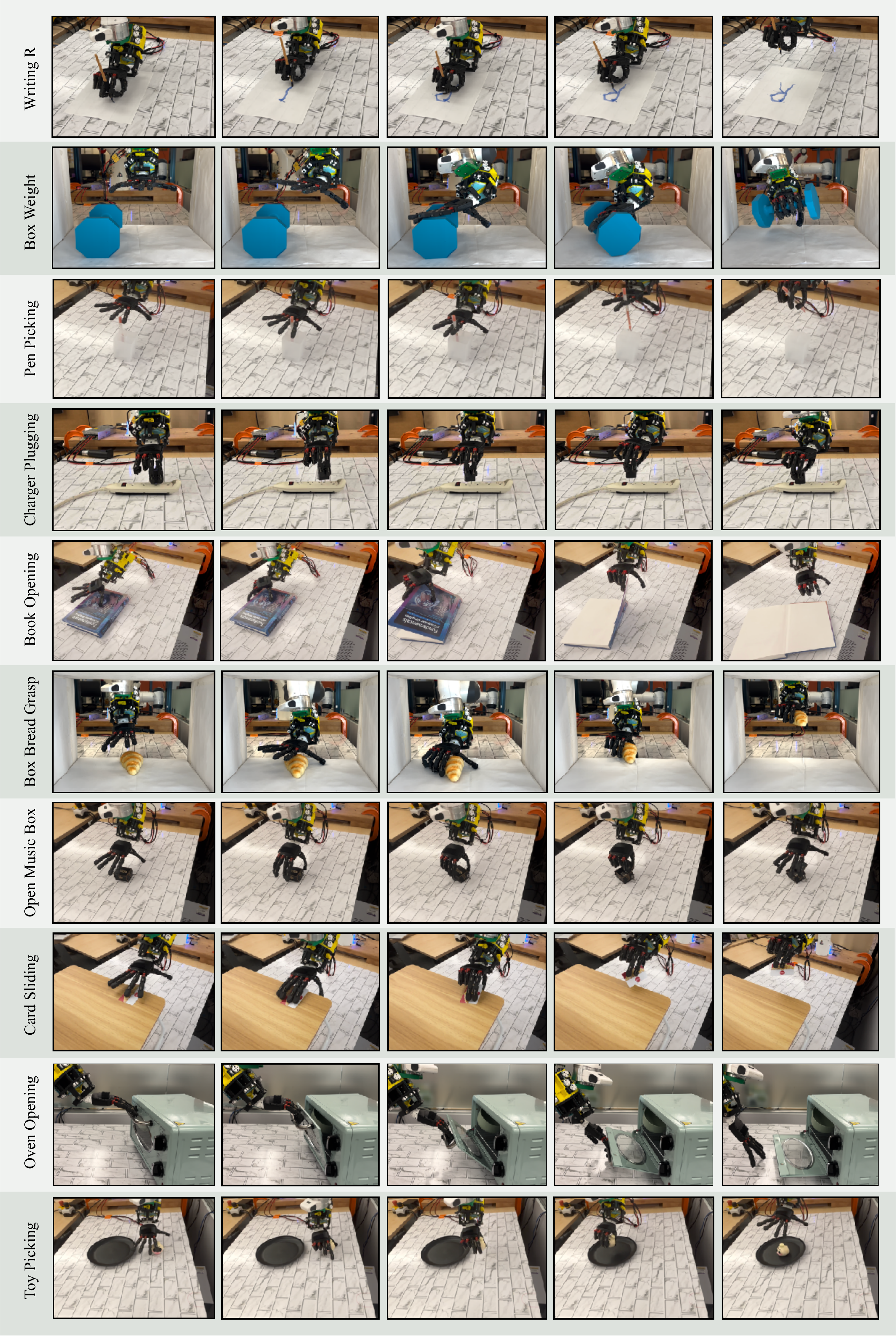}
    \caption{Single arm teleoperated tasks.}
    \label{fig:teleop_figure}
\end{figure}

To perform tasks, we mount \method{} on a 7-DOF Franka arm. We use the Open Teach framework \cite{open-teach} to teleoperate the system with an Oculus VR headset. Wrist trajectories obtained from the Oculus headset are mapped to the Franka arm's end-effector using Open Teach's inverse kinematics pipeline, and hand poses are passed through our controller pipeline to generate \method{} motor commands. Images for teleoperated single arm tasks can be found in \cref{fig:teleop_figure}.

\begin{figure}[H]
    \centering
    \includegraphics[width=\linewidth]{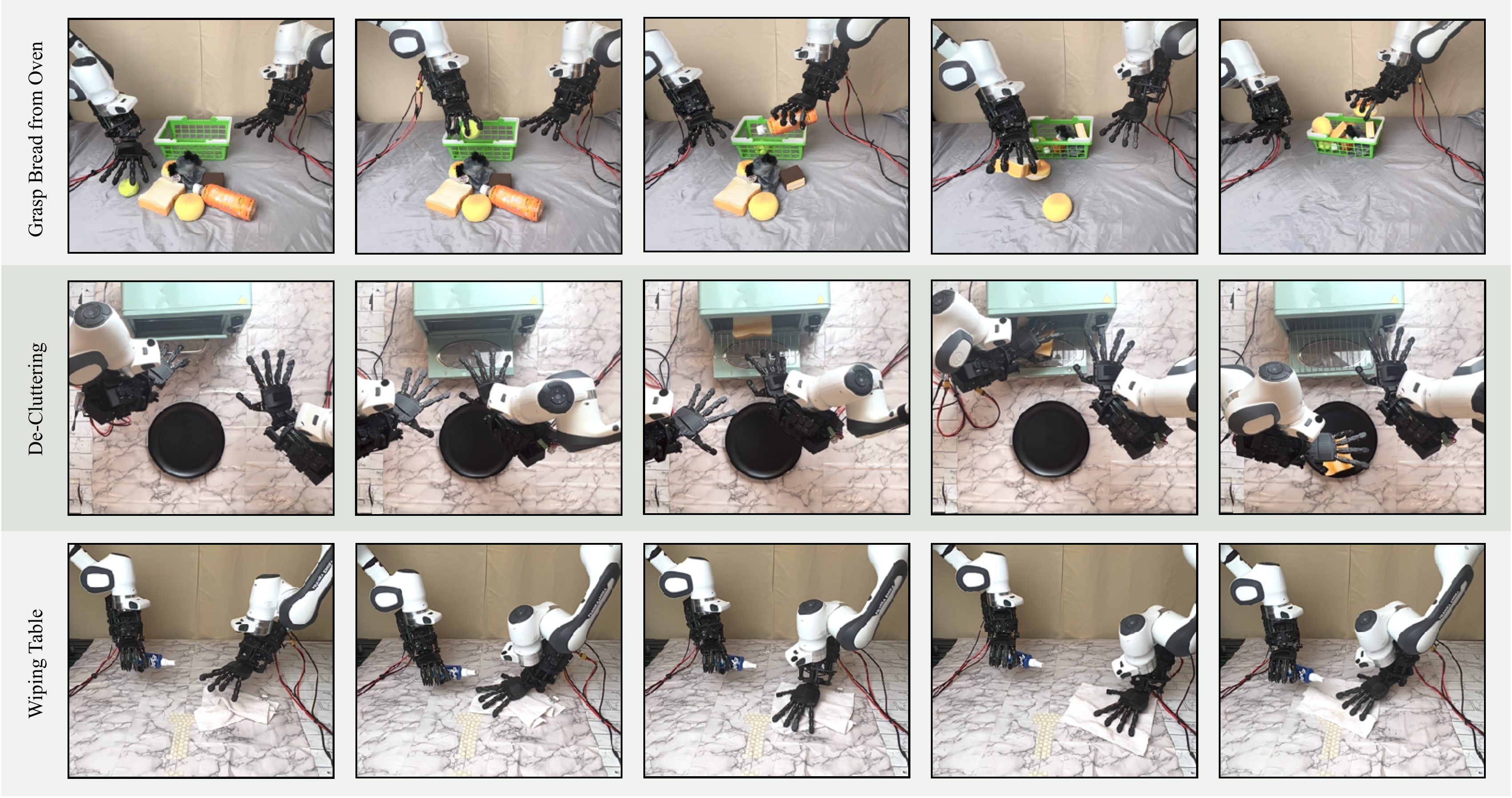}
    \caption{Bimanual teleoperated tasks.
    }
    \label{fig:bim_figure}
\end{figure}

For bimanual tasks, we synchronize two identical Franka arm setups. We perform 10 tasks with a single-arm setup and 3 tasks with a bimanual setup. Images for bimanual tasks can be found in \cref{fig:bim_figure}.

\subsection{Policy Learning}

To evaluate \method{}'s capability in learning complex manipulation tasks, we use the BAKU~\cite{haldar2024baku} framework to train policies for three dexterous tasks: bread pick-and-place, opening a music box, and pen grasping using the abduction/adduction joints of \method{}. We use the OpenTeach framework to collect approximately 100 teleoperated expert demonstrations per task. To improve robustness and error recovery during rollout, following~\cite{visk}, we inject Gaussian noise into the joint states of both the Franka arm and \method{} during data collection. Since operators naturally produce corrective behaviors in response to these perturbations, this automatically enriches the dataset with recovery demonstrations.

Our observation space includes a 23-dimensional proprioception state (7-DOF Franka arm state and the 16-DOF \method{} joint positions) and an RGB video stream from a fixed camera. Using the BAKU architecture, the RGB input is passed through a ResNet-18~\cite{resnet} visual encoder and the proprioception data is encoded with a Multi-Layer Perceptron. These encoded inputs along with a learned action token are then concatenated and fed into a transformer-based~\cite{vaswani2017attention} observation trunk, and the output embedding corresponding to this action token is passed to an action-chunking head to predict a sequence of future actions.

\begin{figure}[H]
    \centering
    \includegraphics[width=\linewidth]{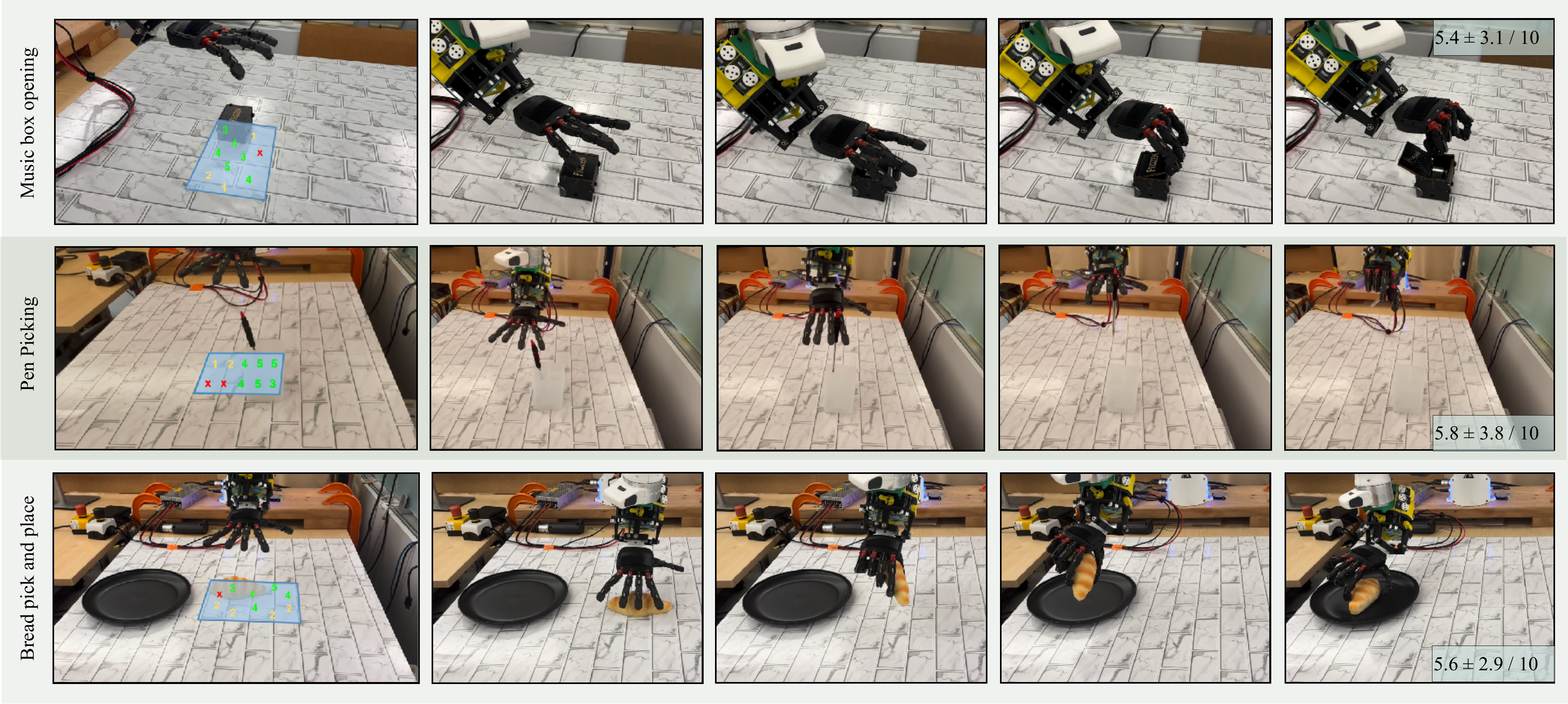}
    \caption{\textbf{Policy rollouts.} Success rates for each starting position are reported out of 5, and total mean and standard deviation success rates are scaled to a 10 point scale. 
    }
    \label{fig:policy_rollouts_figure}
\end{figure}

We evaluate our trained policies at 10 random initial positions per task, across a 20 x 15 cm area for bread pick and place, 15 x 12 cm area for music box opening, and 15 x 10 cm area for pen picking. At each position, we perform 5 closed-loop rollouts and record the mean and standard deviation of the number of successes (out of 5) at each of the 10 test positions. We report the mean and standard deviation scaled to a 10 point scale. Images for the policy rollouts can be found in \cref{fig:policy_rollouts_figure}.

\section{Experimental Evaluation}

We conduct a series of evaluations to assess \method{}'s capabilities as a research tool for robot learning, addressing the following questions:
\begin{enumerate}
\item How does \method{} maintain thermal stability when operated for extended periods?
\item What is the maximum payload \method{} can carry?
\item How accurate is the linear controller mentioned in Section~\ref{sec:controller}? 
\item How does coupling mentioned in Section~\ref{sec:coupling} impact the motion?
\item How does \method{} compare to \ruka{} in teleoperation across a range of tasks?
\end{enumerate}

\subsection{Thermal Endurance}
\label{sec:experiments}

\begin{figure}[t]
    \centering
    \includegraphics[width=\linewidth]{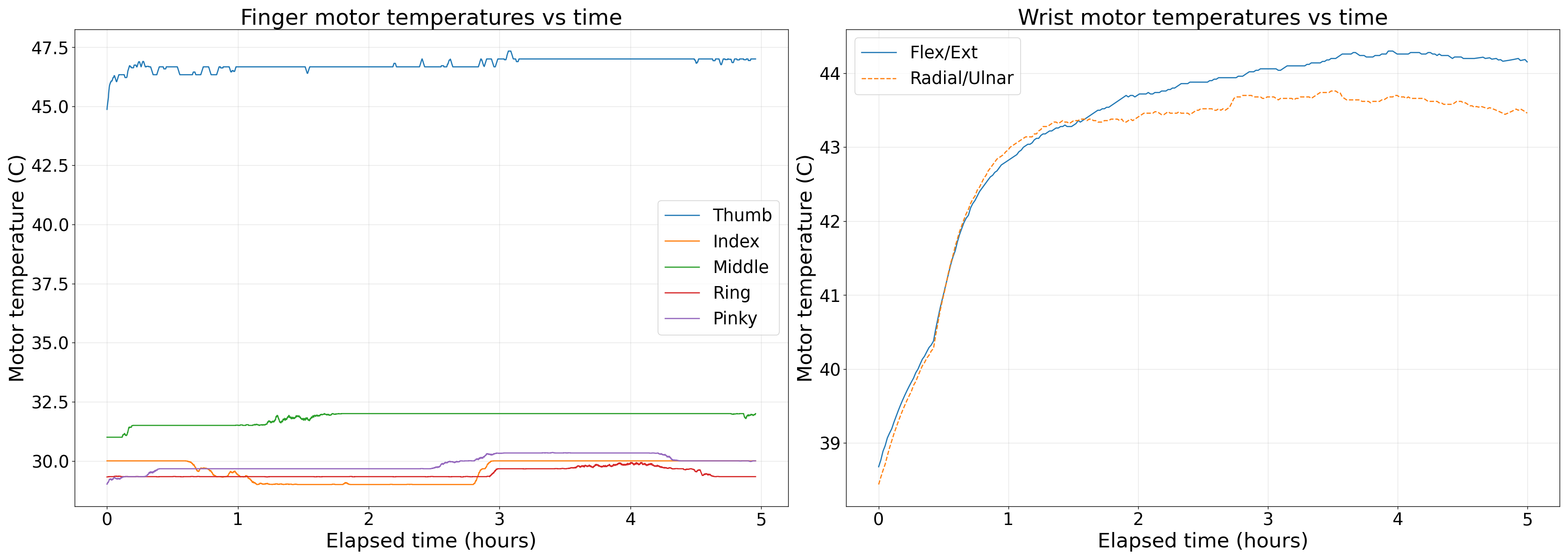}
    \caption{\textbf{5-hour thermal endurance.} Finger motor temperatures (left) and wrist motor temperatures (right) over 5 hours of continuous operation. The wrist motors (Flex/Ext and Radial/Ulnar) gradually rise and stabilize near 44--45\,\textdegree C, while the finger motors remain lower and largely stable throughout, with the thumb motor highest at ${\sim}$47\,\textdegree C. No thermal limiting was observed during the 5-hour run.}
    \label{fig:thermal_5hr}
\end{figure}

To evaluate the thermal reliability of the hand during prolonged operation, we conducted a continuous runtime test over a period of 5 hours. During this experiment, first each degree of freedom for the fingers was actuated and released fully sequentially. This was followed by back-and-forth motion of both DOFs for the wrist joints. Motor temperatures were recorded throughout the test to monitor heat accumulation and identify any risk of overheating or thermal throttling.

Figure~\ref{fig:thermal_5hr} shows the temperature profiles of all actuated fingers over the duration of the experiment, and Table~\ref{tab:thermal_summary} summarizes the corresponding thermal statistics for the finger, thumb, and wrist motor groups. Across all motors, the temperature increased during the initial phase of operation ($\approx 30$ min) and then gradually approached a steady state. No motor exceeded its safe operating limit during the test, and no thermal shutdown or performance degradation was observed.

Among the three motor groups, the finger motors showed the lowest thermal variation, with a peak temperature of 30.75$^\circ$C, a steady-state temperature of 30.35$^\circ$C, and a temperature rise of only 0.81$^\circ$C. The thumb motors reached a higher steady operating temperature, with a peak of 47.33$^\circ$C and a steady-state value of 46.97$^\circ$C, corresponding to a temperature rise of 1.70$^\circ$C. The wrist motors exhibited the largest thermal excursion, reaching a peak temperature of 46.00$^\circ$C, settling to 43.80$^\circ$C at steady state, and showing a total temperature increase of 7.25$^\circ$C.

These results suggest that the \method{} can sustain extended operation without immediate thermal failure. The measured steady-state temperatures also indicate that the tendon routing, transmission design, and actuator selection remain feasible for long-duration manipulation tasks. In particular, although the wrist motors experienced the largest temperature rise, their temperatures remained within a safe operating range throughout the full 5-hour test.

\begin{table}[H]
    \centering
    \caption{Summary of motor temperatures measured during the 5-hour thermal endurance test.}
    \label{tab:thermal_summary}
    \begin{tabular}{lccc}
        \toprule
        \textbf{Motor group} & \textbf{Peak ($^\circ$C)} & \textbf{Steady-state ($^\circ$C)} & \textbf{$\Delta T$ ($^\circ$C)} \\
        \midrule
        Fingers & 30.75 & 30.35 & 0.81 \\
        Thumb   & 47.33 & 46.97 & 1.70 \\
        Wrist   & 46.00 & 43.80 & 7.25 \\
        \bottomrule
    \end{tabular}
\end{table}

\subsection{Payload}
We evaluated the payload capability of the hand by measuring the maximum load that could be statically supported for a fixed duration under different joints and wrist orientations. In each trial, a bag was attached to the tested finger joint or placed on the palm, and weights were gradually added until the hand could no longer maintain the target posture. The maximum load was defined as the largest weight that could be held stably for the specified hold time without visible failure, such as joint collapse, tendon slip, or loss of posture.
\begin{figure}[t]
    \centering
    \begin{minipage}[t]{0.327\linewidth}
        \centering
        \includegraphics[height=5.2cm, width=1.25\linewidth,
            , angle=270, origin=c]{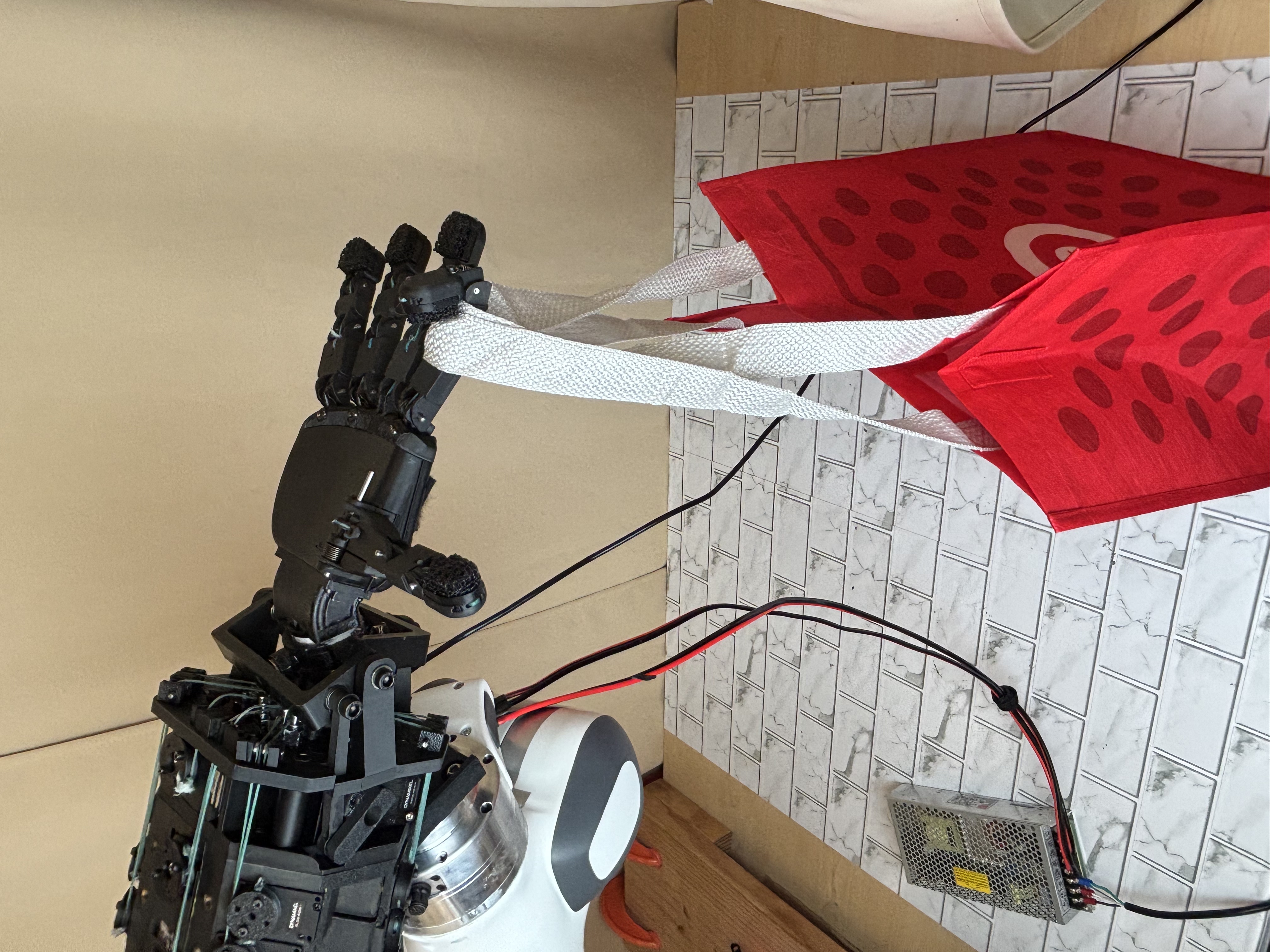}
        \vspace{2pt}
        \small \vspace{2pt}  (a) Finger DIP--PIP load
    \end{minipage}
    \hfill
    \begin{minipage}[t]{0.327\linewidth}
        \centering
        \includegraphics[height=5.2cm, width=1.25\linewidth,
            , angle=270, origin=c]{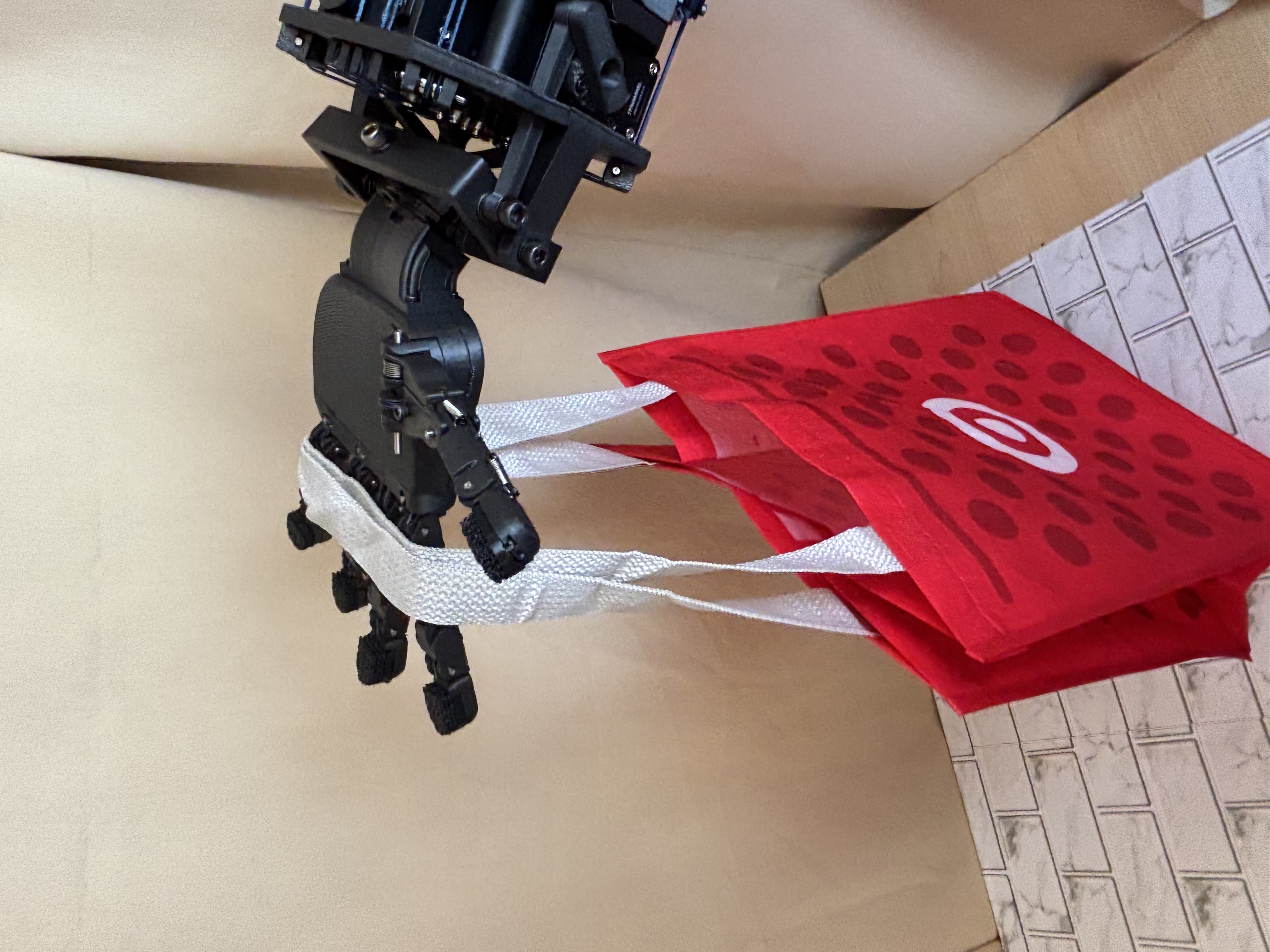}
        \vspace{2pt}
        \small \vspace{2pt}  (b) Wrist supination
    \end{minipage}
    \hfill
    \begin{minipage}[t]{0.327\linewidth}
        \centering
        \includegraphics[height=5.2cm, width=1.25\linewidth,
        , angle=270, origin=c]{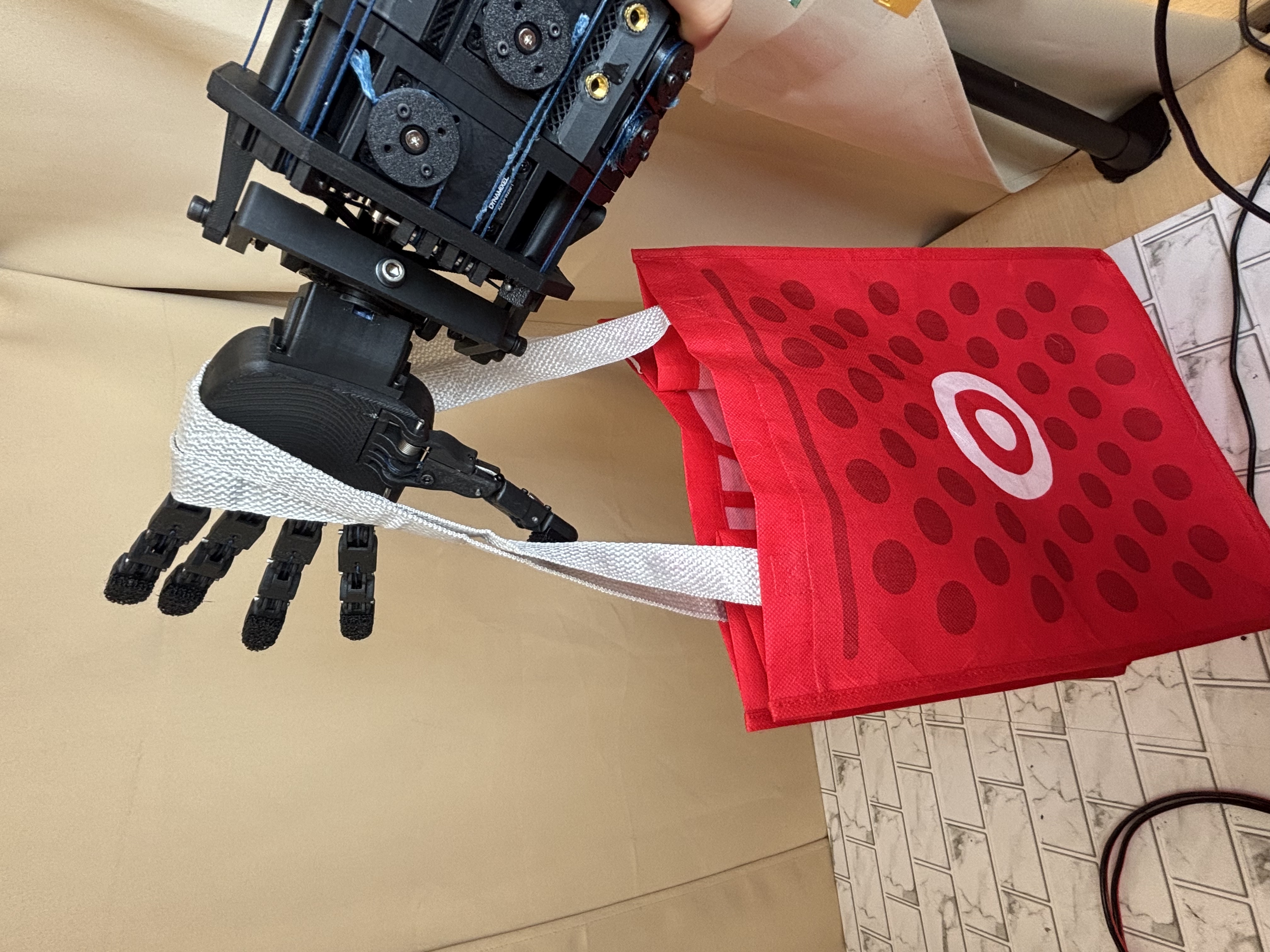}
        \vspace{2pt}
        \small  \vspace{2pt} (c) Wrist radial/ulnar
    \end{minipage}
    \caption{\textbf{Finger load test setup.} Static payload evaluation under representative
    loading conditions: finger DIP--PIP joint load (\textbf{a}), wrist in forearm supination
    with palm face-up (\textbf{b}), and wrist in sideways (radial/ulnar-side-up) orientation
    (\textbf{c}). In each trial, a bag with incrementally added weights was suspended from the
    hand, and the maximum load held stably for the specified duration was recorded.}
    \label{fig:payload_setup}
\end{figure}
\begin{table}[t]
    \centering
    \caption{Static payload performance of the hand under different joint and wrist loading conditions.}
    \label{tab:payload_summary}
    \begin{tabular}{lcc}
        \toprule
        \textbf{Condition} & \textbf{Max load (g)} & \textbf{Hold time (s)} \\
        \midrule
        Non-thumb fingers DIP-PIP & 1200 & 15 \\
        Non-thumb fingers MCP & 780 & 15 \\
        Non-thumb fingers adduction & 150 & 15 \\
        Thumb & 835 & 20 \\
        Wrist (forearm supination) & 1215 & 20 \\
        Wrist (forearm pronation) & 1215 & 20 \\
        Wrist (radial-side-up) & 835 & 20 \\
        Wrist (ulnar-side-up) & 835 & 20 \\
        \bottomrule
    \end{tabular}
\end{table}
Table~\ref{tab:payload_summary} summarizes the measured payloads. For the non-thumb fingers, the coupled DIP--PIP joints supported the highest load of 1200\,g for 15\,s, while the MCP joint supported 780\,g for the same duration. The non-thumb finger adduction configuration supported a substantially smaller load of 150\,g, reflecting the lower stiffness and moment capacity of this motion direction. The thumb was evaluated by curling all three joints and suspending a weighted bag from the finger; weights were incrementally added until any joint began to move, with the maximum stable load recorded as 835\,g for 20\,s.

We also evaluated wrist payload under four representative forearm orientations: supination, pronation, radial-side-up, and ulnar-side-up. The wrist supported 1215\,g for 20\,s in both the forearm supination and pronation configurations, and 835\,g for 20\,s in both the radial-side-up and ulnar-side-up configurations. This result indicates that wrist load capacity is orientation-dependent, with greater load tolerance when the applied moment is aligned with the stronger structural direction of the mechanism.
Overall, these results show that the hand can sustain substantial static loads across multiple joints and orientations. The payload differences across configurations also highlight the anisotropic mechanical behavior of the tendon-driven structure, which should be considered in future task planning.

\subsection{Controller Accuracy}
\label{sec:controller_accuracy}

To evaluate the accuracy of the controller, we employed the attachable magnetic encoders in Section \ref{sec:magnetic_encoder} to obtain ground truth joint angle measurements $\theta_{actual}$. Given that the magnetic sensors' initial readings are dependent on the orientation of the poles once mounted on the joints, the ground truth joint angles were shifted to match the minimum angle for each joint. 

The test evaluates each of the 4 index finger joints and 3 thumb joints individually by sampling 20 random angles $\theta_{expected}$ within the joint limits of each joint and calculating the motor commands $p$ corresponding to these angles using the equation described in Section \ref{sec:controller}. The motors are moved using the motor commands $p$, and each resulting position is held for 1.5 seconds. We evaluate control accuracy by calculating the absolute error between the commanded angles $\theta_{expected}$ and ground truth readings $\theta_{actual}$. To provide a standardized metric across varying ranges of motion, we also compute the normalized error expressed as a percentage of the joint range $[\theta_{min}, \theta_{max}]$.

\begin{figure}[H]
    \centering
    \includegraphics[height=7cm, width=\linewidth]{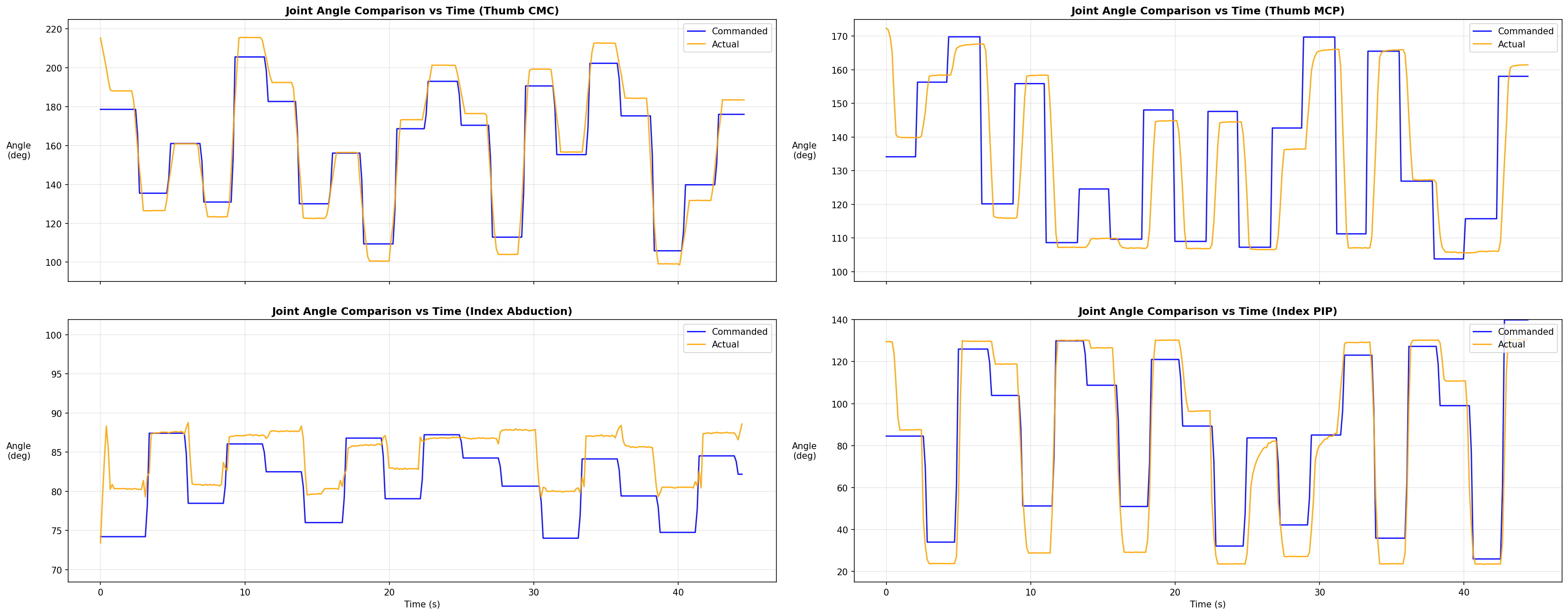}
    \caption{\textbf{Joint Angle Comparison.} Comparison between expected angles $\theta_{expected}$ and ground truth angles measured by the encoder.}
    \label{fig:joint_angle_graph}
\end{figure}

We observe a total average error of $8.26^\circ$ and total average percentage error of $10.68\%$. Results can be found in Figure \ref{fig:joint_angle_graph} and Table \ref{tab:joint_error_summary}. 

\begin{table}[H]
    \centering
    \caption{Average joint angle tracking error for the \method{} hand during individual joint movement phases.}
    \label{tab:joint_error_summary}
    \begin{tabular}{lcc}
        \toprule
        \textbf{Joint Name} & \textbf{Avg. Error ($^\circ$)} & \textbf{Avg. Error (\%)}\\
        \midrule
        Index Abduction & 3.76 & 15.96\% \\
        Index DIP & 7.69 & 9.62\% \\
        Index PIP & 11.91 & 11.91\% \\
        Index MCP & 12.77 & 8.51\% \\
        Thumb CMC & 7.85 & 7.47\% \\
        Thumb MCP & 11.07 & 16.27\% \\
        Thumb IP & 2.76 & 5.01\% \\
        \midrule
        \textbf{Overall Average} & \textbf{8.26$^\circ$} & \textbf{10.68\%}\\
        \bottomrule
    \end{tabular}
\end{table}

\subsection{Difference in Motion with Coupled DIP/PIP}

To evaluate the effect of the redesigned DIP--PIP coupling mechanism, we compared the measured joint trajectories of the previous finger design and the coupled \method{} finger under repeated flexion--extension sweeps. The index finger was driven through repeated motion cycles while the joint angle was measured against the commanded motor position. The comparison focuses on motion repeatability, spread across repeated sweeps, and the consistency of the command-to-angle relationship.

\begin{figure}[H]
    \centering
    \includegraphics[height=5.8cm,width=\linewidth]{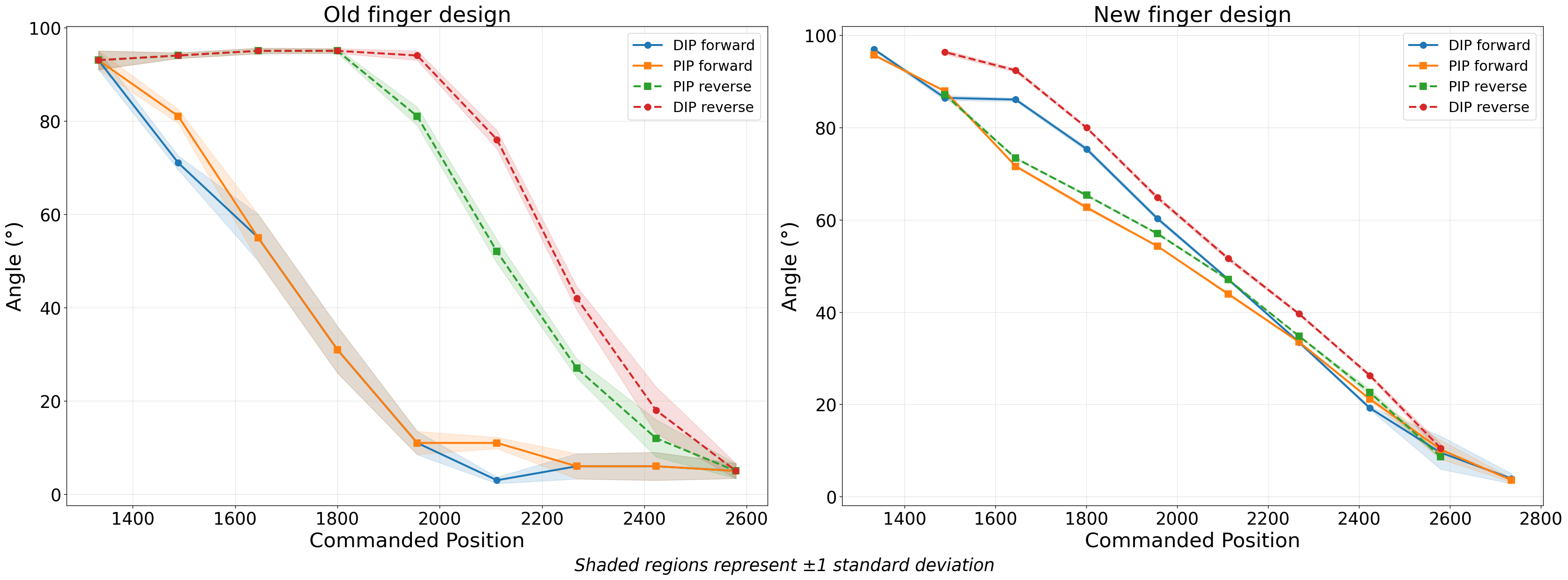}
    \caption{\textbf{Effect of DIP--PIP coupling on motion repeatability.} Angle versus commanded position for repeated flexion--extension sweeps, comparing the previous finger design (left) and the redesigned coupled DIP--PIP mechanism (right). Solid lines show forward motion; dashed lines show reverse. Shaded regions represent $\pm1$ standard deviation across trials. The redesigned mechanism exhibits significantly reduced hysteresis and trial-to-trial variation, yielding a more consistent command-to-angle relationship across the full range of motion.}
    \label{fig:dip_pip_compare}
\end{figure}

Figure~\ref{fig:dip_pip_compare} compares the angle-versus-commanded-position trajectories for the previous design and the redesigned coupled mechanism. In the previous design, the trajectories exhibit larger trial-to-trial variation, especially near the lower-angle region and during direction reversal. This indicates stronger effects of tendon slack, friction, and small mechanical inconsistencies across repeated sweeps.

In contrast, the redesigned coupled mechanism produces a more stable and repeatable trajectory. The measured curves are more tightly clustered, and the overall relationship between commanded position and resulting joint angle is more consistent across trials. This improvement is attributed to the addition of the fixed-length DIP--PIP coupling strings, which better constrain the geometric relationship between the middle and distal phalanges and reduce variability caused by slack and friction in the original design.

Overall, the coupled DIP--PIP design improves motion consistency and makes the finger behavior more predictable under repeated actuation. This increased repeatability is beneficial for both grasp execution and controller design, since a simpler calibration-based mapping can more reliably reproduce desired finger postures.

\subsection{\ruka{} / \method{} User Experience Comparisons}
\label{sec:v1-comparison}

To evaluate the usability of \method{} compared to the RUKA, a user experience study was conducted with 10 participants from our lab. Each participant performed three single-arm manipulation tasks using the teleoperation setup described in Section~\ref{sec:teleoperation}. Participants were given a two-minute period to familiarize themselves with the teleoperation framework. Then, each participant performed three trials per task. Completion times were recorded in seconds, with a maximum time limit of 120 seconds; trials exceeding 120 seconds were counted as failures.

The evaluation consisted of the following tasks:
\begin{enumerate}
    \item \textbf{Bread Pick-and-Place: }Picking up a bread item from the table and placing it onto a plate.
    \item \textbf{Pen Grasping: }Picking up a pen from a vertical pen holder using the hand’s abduction/adduction joints.
    \item \textbf{Book Opening: }Opening the cover of a hard-cover book placed flat on a table.
\end{enumerate}

\begin{figure}[H]
    \centering
    \begin{minipage}[t]{0.487\linewidth}
        \centering
        \includegraphics[width=\linewidth]{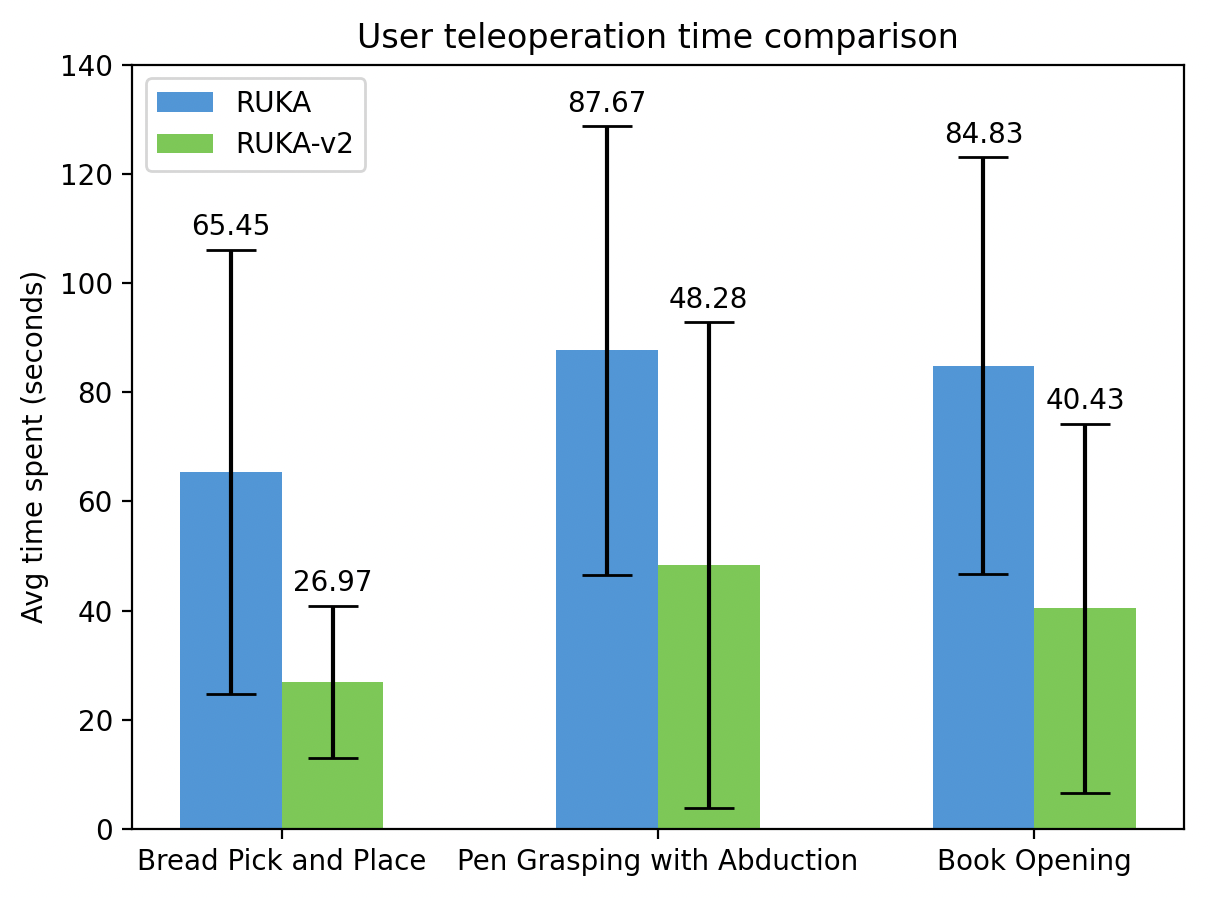}
        
        \vspace{2mm}
        \small (a) Teleoperation time comparison
    \end{minipage}
    \hfill
    \begin{minipage}[t]{0.48\linewidth}
        \centering
        \includegraphics[width=\linewidth]{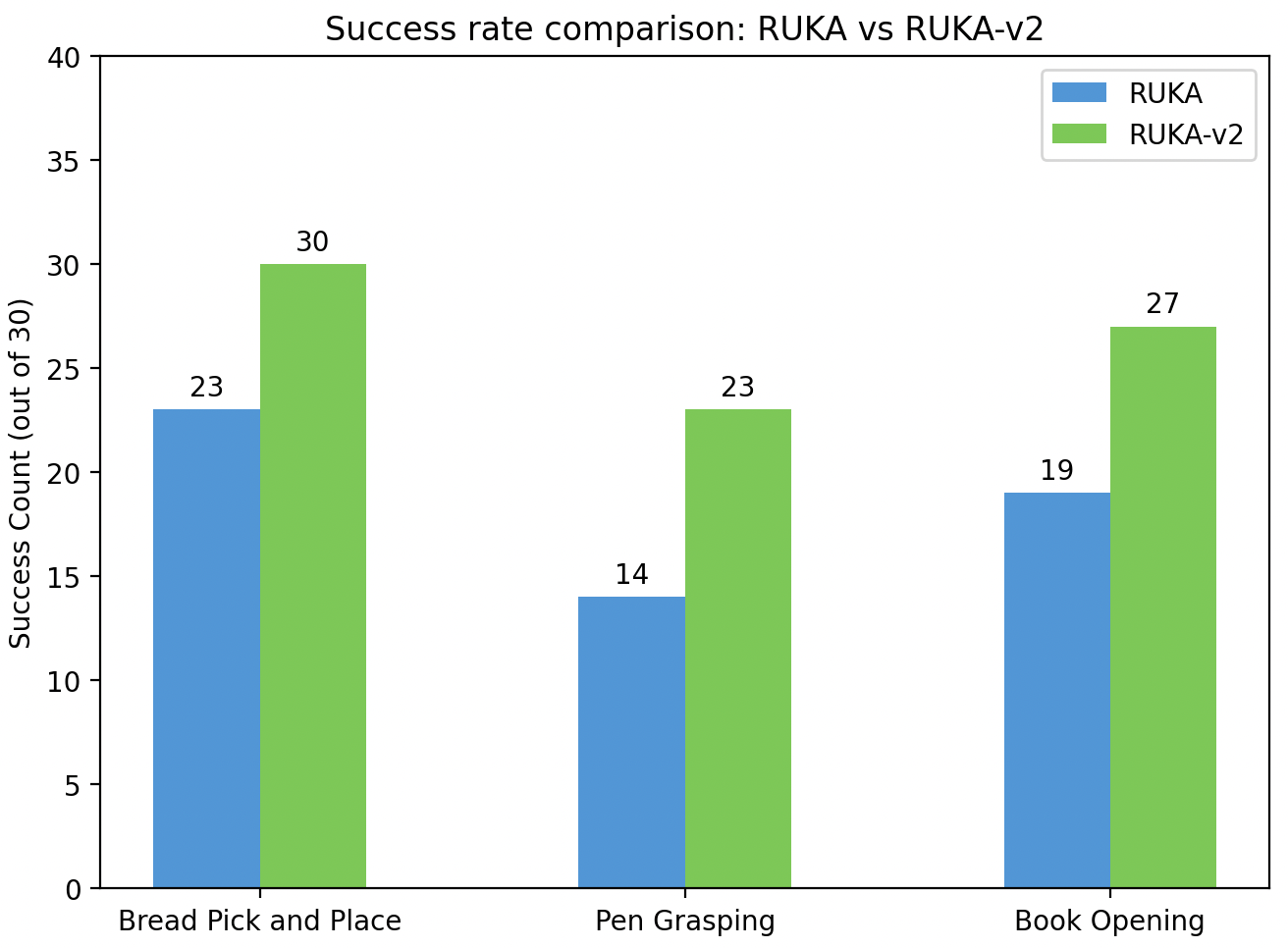}
        
        \vspace{2mm}
        \small (b) Success rate comparison
    \end{minipage}
    \caption{\textbf{\ruka{}/\method{} user experience comparisons} We
    observed a 51.3\% reduction in mean completion time and a 21.2\% increase in overall success rates across all tasks.}
    \label{fig:v1-comparison}
\end{figure}

The transition to \method{} resulted in significant performance improvements across all metrics. We observed a 51.3\% reduction in mean completion time and a 21.2\% increase in overall success rates compared to RUKA. Results of teleoperation times and success rates per task can be found in \cref{fig:v1-comparison}.


\section{Discussion}
In this work, we presented \method{}, an improvement over \ruka{} that adds a 2-DOF decoupled parallel wrist and per-finger adduction/abduction. While we demonstrate how these additions yield more capable hardware, we acknowledge the following limitations and directions for future work:

\paragraph{Assumption of linear joint-to-motor mapping} Similar to~\cite{orca} and unlike \ruka{}, we use linear interpolation between motor and joint limits to compute the motor positions required for a desired joint angle, implicitly assuming a linear relationship between the two. While this yields sufficiently accurate control (Section~\ref{sec:controller_accuracy}), the linearity assumption has not been formally verified. \ruka{} introduced a data-driven approach to model this relationship; a similar approach could be applied here using the magnetic sensors introduced in this work, replacing the need for expensive motion capture gloves while maintaining accurate control.

\paragraph{Out-of-the-box tactile integration} While we introduce e-flesh fingertips that can be converted into tactile sensors, adding touch sensing to a multi-fingered hand may introduce noise due to magnetic interference between sensors. Future work should characterize the extent of this interference and work toward seamless out-of-the-box tactile integration. Potential directions include optimizing magnet placement within the e-flesh to minimize interference, or training a data-driven model to identify and compensate for it.


\section*{Acknowledgments}
We thank Billy Yan, Anya Zorin, Zichen Cui, Kevin Wu, Yihang Zhou, Kelly Lee and Alex Jiang for helps on user experiments, valuable feedback, and discussions.
\paragraph*{Funding:} This work was supported by grants from LG, Qualcomm, Honda, Microsoft, NSF award 2339096, and ONR award N00014-22-1-2773. Lerrel Pinto is supported by the Sloan and Packard Fellowships.
\paragraph*{Author contributions:}
Xinqi and Ruoxi co-led the project. Xinqi designed the wrist and finger adduction mechanisms and conducted hardware experiments with guidance from Irmak, Raunaq and Lerrel. Ruoxi implemented the software stack, collected demonstrations and trained policies with \method{} with the help of Zhuoran and Raunaq. Alejandro designed the magnetic encoders and conducted accuracy experiments with them. Ruoxi and Irmak conducted the user experience experiments. Kenny and Charles helped build the bimanual setup. Irmak led the writing of the manuscript with help from Ruoxi, Xinqi and Alejandro. All authors provided feedback and revisions. 
\paragraph*{Competing interests:}
There are no competing interests to declare.
\paragraph*{Data and materials availability:}
All data and materials needed to reproduce and evaluate the results and conclusions of this manuscript are present in the paper or in the Supplementary Materials.

\bibliography{references}


\end{document}